\documentclass{SCIS2025}
\usepackage{times}
\usepackage{helvet}
\usepackage{courier}
\usepackage{graphicx}
\usepackage{textcomp}
\usepackage{xcolor}
\usepackage{mdframed}
\usepackage{multirow}
\usepackage{array}
\usepackage{makecell}
\usepackage{booktabs}
\usepackage{stackengine}
\usepackage{multicol}
\usepackage{tabularx}
\usepackage{ragged2e}
\usepackage{balance}
\usepackage{cite}
\usepackage{hyperref}

\usepackage{subcaption}
\begin{document}
\ArticleType{RESEARCH PAPER}
\Year{2025}
\Month{}
\Vol{}
\No{}
\DOI{}
\ArtNo{}
\ReceiveDate{}
\ReviseDate{}
\AcceptDate{}
\OnlineDate{}
\AuthorMark{}
\AuthorCitation{}

\title{Graph Neural Architecture Search with Large Language Models}{Graph Neural Architecture Search with Large Language Models}

\author[1*\textdagger]{Haishuai Wang}{{haishuai.wang@zju.edu.cn}}
\author[2\textdagger]{Yang Gao}{}
\author[1\textdagger]{Xin Zheng}{}
\author[3*]{Peng Zhang}{{p.zhang@gzhu.edu.cn}}
\author[1]{Jiajun Bu}{}
\author[4]{Philip S. Yu}{}

\contributions{Haishuai Wang, Yang Gao and Xin Zheng contributed equally to this work.}


\address[1]{Zhejiang Key Laboratory of Accessible Perception and Intelligent Systems, \\ College of Computer Science and Technology, Zhejiang University, Hangzhou, 310027, China}
\address[2]{School of Public Health, Zhejiang University, Hangzhou 310058, China}
\address[3]{Cyberspace Institute of Advanced Technology, Guangzhou University, Guangzhou 510006, China}
\address[4]{Department of Computer Science, University of Illinois at Chicago, Chicago 60607, USA}
\abstract{Graph Neural Architecture Search (GNAS) has shown promising results in finding the best graph neural network architecture on a given graph dataset. However, existing GNAS methods still require intensive human labor and rich domain knowledge when designing the search space and search strategy. To this end, we integrate Large Language Models (LLMs) into GNAS and present a new GNAS model based on LLMs (\textbf{GNAS-LLM} for short). The basic idea of GNAS-LLM is to design a new class of \textit{GNAS prompts} for LLMs to guide LLMs towards understanding the generative task of graph neural architectures. The prompts consist of descriptions of the search space, search strategy, and search feedback of GNAS. By iteratively running LLMs with the prompts, GNAS-LLM generates more accurate graph neural network architectures with fast convergence. Experimental results show that GNAS-LLM outperforms the state-of-the-art GNAS methods on four benchmark graph datasets, with an average improvement of 0.7\% on the validation sets and 0.3\% on the test sets. Besides, GNAS-LLM achieves an average improvement of 1.0\% on the test sets based on the search space from AutoGEL.}


\maketitle

\section{Introduction}
 
The rapid growth of graph data has propelled the development of graph neural networks (GNNs) to capture data dependencies between graph nodes. Today, GNNs have been popularly used as an effective tool for a wide range of graph data  applications~\cite{wang2016incremental,gaoprecision,chen2024deepasd,li2021live}. However, given a graph dataset, designing the best GNN architecture is a challenging task which heavily relies on manual exploration and domain expertise. This is because the neural architecture space of GNNs is vast and diverse, with numerous layer configurations and connectivity patterns.

Recently, graph neural architecture search (GNAS) ~\cite{DBLP:conf/ijcai/ZhangW021} has emerged as a promising solution for automatically designing GNN architectures. The base idea of GNAS is to manually design a graph neural architecture search space, based on which a search strategy is developed to explore the search space with respect to a given evaluation metric. As a result, the best architecture on a validation graph dataset is returned as the solution of GNAS. Existing GNAS methods can be categorized into three types according to their search strategies, i.e., reinforcement learning GNAS~\cite{DBLP:conf/ijcai/GaoYZ0H20}, differential gradient GNAS~\cite{DBLP:conf/icde/ZhaoYT21}, and evolutionary GNAS~\cite{Shi2020EvolutionaryAS}.
Although these GNAS methods have obtained commendable results, the design of graph neural architecture search algorithms requires heavy manual work with domain knowledge.

Recent advances in natural language processing have introduced a series of powerful large language models (LLMs) that have shown remarkable language understanding and generation capability. More importantly, LLMs have been used to design new neural architectures for CNNs~\cite{zheng2023can}. The key idea is to design a new class of prompts to guide LLMs to pinpoint promising CNNs and learn from historical attempts. 
Inspired by the work, an intuitive question arises as follows: \textit{Can we make use of the powerful generation capability of LLMs to generate new graph neural architectures, so as to alleviate the burden of manually designing the search space and search strategy of GNAS?} 
Generally, GNAS operations are different from traditional CNN operations, particularly due to their irregular message passing and aggregation functions, which leads to a more complicated and diverse GNN search space~\cite{DBLP:conf/ijcai/GaoYZ0H20,DBLP:conf/cvpr/Cai0DZZ0H21,zhang2023autogt}, and thus the application of large language models to GNAS is very challenging.

This paper aims to answer the above question and explore the utilization of LLMs for GNAS to generate new graph neural architectures.  The core idea is to design new class of \textit{GNAS prompts} for LLMs which can leverage the generation capability of LLMs to generate new GNN architectures. Specifically, we present a new LLMs based Graph Neural Architecture Search method (\textbf{GNAS-LLM} for short) which introduces new prompts to guide LLMs towards understanding the search space and search strategy of GNAS.
First, GNAS-LLM takes LLMs as controller to  generate new GNN architectures by iteratively exploring the search space. Then, GNAS-LLM uses the evaluation results of the generated GNN architectures as rewards to improve the GNAS prompts. After a few iterations, GNAS-LLM  converges very fast to the best GNN architectures. The contributions of this work are summarized as follows: 

 \begin{itemize}
 
\item This represents the first effort to jointly studying graph neural architectures and large language models, where a new model GNAS-LLM is proposed to embed LLMs into GNAS by taking LLMs as the  controller of GNAS. 

\item A new class of \textit{GNAS prompts}  is designed for LLMs. These prompts can guide LLMs to understand the search space and search strategy of GNAS. To understand the search space, the prompts use an adjacency matrix to describe the space with both candidate operations and candidate connections. To understand the search strategy, the prompts include descriptions of reinforcement learning with both exploration and exploitation. 

\item Experimental results indicate that GNAS-LLM can generate better architectures than existing GNAS methods with less search iterations, e.g., GNAS-LLM reduces  56\% iterations on average on the test sets. The codes of this work are released on Github: \url{https://github.com/checkuredu/GNAS-LLM}. 

 \end{itemize}

Compared with the previous LLM-enhanced Neural Architecture Search (NAS) method ~\cite{zheng2023can} designed for only one  search space with a fixed search strategy, GNAS-LLM can adapt to multiple search spaces and multiple search strategies. 
We summarize the technical progress from three aspects. First, our approach is scalable to new GNN operations, where the search space can be easily expanded by adding new operations to the GNAS prompts. 
Second, we describe GNN connections between operations in the GNAS prompts, which enables GNAS-LLM adaptable to new search spaces conveniently.
Third, our approach is adaptable to new  search strategies by only revising the search strategy prompt. 

\section{Related Works}
\subsection{Graph Neural Architecture Search (GNAS)} 

GraphNAS~\cite{DBLP:conf/ijcai/GaoYZ0H20} represents an early effort that uses reinforcement learning to design  GNN architectures. Based on GraphNAS, AutoGNN~\cite{AutoGNN} introduces an entropy-driven candidate model sampling method and a new weight-sharing strategy to efficiently select GNN components.  GraphNAS++~\cite{gao2022graphnas++} uses distributed architecture evaluation to accelerate GraphNAS. GM2NAS~\cite{gao2023gm2nas} uses reinforcement learning to design GNNs for multitask multiview graph learning. MVGNAS~\cite{al2022multi} is designed for biomedical entity and relation extraction. Besides, HGNAS~\cite{DBLP:conf/icdm/GaoZLZLH21} and HGNAS++~\cite{gao2023hgnas++} use  reinforcement learning to find  heterogeneous graph neural networks.

Different from reinforcement learning based GNAS that explores a discrete GNN search space, a new class of differentiable gradient GNAS methods were proposed to explore a relaxed yet continuous GNN search space, such as  DSS~\cite{li2021one}, SANE~\cite{DBLP:conf/icde/ZhaoYT21}, GAUSS~\cite{guan2022large}, GRACES~\cite{qin2022graph},  AutoGT~\cite{zhang2023autogt}, and Auto-HEG~\cite{zheng2023auto}. SANE focuses on searching for data-specific neighborhood aggregation architectures. DSS is designed for GNN architecture with a dynamic search space. GAUSS~\cite{guan2022large} addresses large-scale graphs by devising a lightweight supernet and employing joint architecture-graph sampling for efficient handling. GRACES~\cite{qin2022graph} achieves generalization under distribution shifts by adapting a tailored GNN architecture specifically designed for each graph instance with an unknown distribution. AutoGT~\cite{zhang2023autogt} introduces an automated neural architecture search framework that extends GNAS to Graph Transformers. Auto-HeG~\cite{zheng2023auto} enables automatic search for the neural architecture of GNNs for heterophilic graphs. 
In addition, AutoGEL~\cite{wang2021autogel}, DiffMG~\cite{diffmg}, MR-GNAS~\cite{Zheng2022MultiRelationalGN}, DHGAS~\cite{Zhang_Zhang_Wang_Qin_Qin_Zhu_2023}, and HLWP~\cite{zhang2024meta} focus on heterogeneous graphs. 

Recently, AutoGraph~\cite{li2020autograph} and  Genetic-GNN~\cite{Shi2020EvolutionaryAS} use evolutionary algorithms to find the best GNN architectures. G-RNA~\cite{DBLP:conf/cvpr/XieCZWWZY023} proposed a unique search space and define a robustness metric to guide the search procedure in order to search for defensive GNNs.
Zhang et. al~\cite{DBLP:conf/ijcai/ZhangW021} and Oloulade et. al~\cite{oloulade2021graph} survey automatic machine learning methods on graphs. 
However, these works have not touched the problem of using large language models to enhance the GNAS models, which is the focus of this paper. 

\subsection{Large Language Models (LLMs)}
     GPT-4~\cite{openai2023gpt4} represents  a new generation of AI models that can generate answers with respect to questions on multi-modal data. In particular, recent efforts have shown that  GPT-4 is capable of understanding  graph data~\cite{guo2023gpt4graph} and performs well on various graph learning tasks. Moreover, a recent work~\cite{zhanggraph} combines large language models and graph learning models and uses GPT-4 to reason over graph data.  Different from the GPT models, BERT~\cite{DBLP:conf/naacl/DevlinCLT19}  pre-trains a model on large-scale unlabeled data, and then fine-tunes the model on specific downstream tasks. Based on BERT, a number of  language models  were proposed ~\cite{Lan2020ALBERT,DBLP:conf/iclr/HeLGC21}.
    For example, PaLM~\cite{chowdhery2022palm} is built upon the decoder of  Transformers~\cite{Vaswani2017AttentionIA}. PaLM 2~\cite{anil2023palm} uses a larger dataset and a more complicated architecture and obtains better results than PaLM. 
    
The integration of graph learning with large language models (LLMs) has recently attracted extensive attention~\cite{DBLP:journals/corr/abs-2311-12399}. 
TAPE~\cite{he2023harnessing} uses LLMs to generate explanations and pseudo-labels to augment textual attributes of graphs, wherein LLMs act as a data enhancer. 
GraphGPT~\cite{tang2024graphgptgraphinstructiontuning} aligns LLMs with graph structural knowledge through graph instruction tuning. GraphPrompter~\cite{canwesoftpromptllms} consists of two main components, i.e., a graph neural network that encodes complex graph information, and an LLM that effectively processes textual information. Both GraphGPT and GraphPrompter take LLMs as a predictor which directly output results.

    Recently, large language models have been used for neural architecture search. A pioneering work GENIUS~\cite{zheng2023can} uses GPT-4 to design neural architectures for CNNs, aiming to explore the generation capability of LLMs. The key idea is to allow GPT-4 to learn from the feedback of generated neural architectures and iteratively generate better ones. Experimental results on benchmarks demonstrate that GPT-4 can find top-ranked architectures after several iterations of prompts. Subsequently, AutoML-GPT~\cite{zhang2023automl} designs a series of prompts for LLMs to automatically complete tasks such as data processing, model architecture design, and hyper-parameter tuning.
    
    Although the above works are successful, LLMs have not been used to generate graph neural architectures. Therefore, in this paper, we extend LLMs to the task of generating new graph neural architectures. This approach can alleviate the burden of manually designing the search space and search strategy of GNAS and improve the performance of existing GNAS methods.
       
\section{Methods}

In this section, we introduce LLMs for GNAS, which consists of a new class of GNAS prompts that can instruct LLMs to generate new GNNs. Formally, given an LLM model, a dataset $\mathcal{G}$, a GNN search space  $\mathcal{M}$, and an evaluation metric $\mathcal{A}$, we aim to find the best architecture $m^{*} \in \mathcal{M}$ on a given graph  $\mathcal{G}$, i.e., 
\begin{equation}
    m^*=\underset{m\in \mathcal{M}(\textbf{LLM})}{argmax} \ \ \mathcal{A}(m(\mathcal{G}))
\end{equation} 
where $\mathcal{M}(\textbf{LLM})$ denotes the search space generated by the LLM, and the metric $\mathcal{A}$ can be either accuracy or AUC for graph node classification tasks. 

\begin{figure}[t]
      \centering
      \includegraphics[width=1\linewidth]{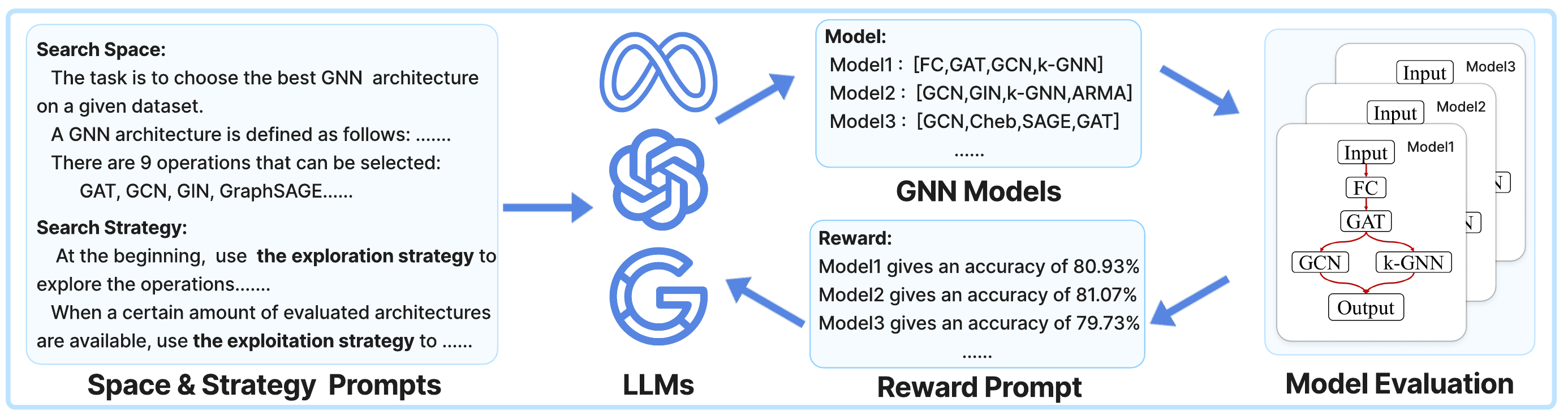}
      \caption{An overview of GNAS-LLM. First,  \textit{GNAS prompts} are designed to describe the search task, search space, and search strategy of GNAS. Then, the GNAS prompts guide LLMs towards generating new  architectures within the search space. Based on the generated  architectures, new GNN models are trained on a given graph and tested  according to a metric, such as accuracy. The generated GNN models and their tested results are returned to LLMs as rewards described by the reward prompts. Repeatedly, LLMs update the GNN models and eventually output the best one.}
      \label{fig:model_fig}
    \end{figure}

\subsection{GNAS Prompts}

To solve Eq. (1), an essential question is to design a new class of prompts that guides LLMs to generate new \textit{candidate GNN architectures}, i.e., $ \mathcal{M}(\textbf{LLM})$. The design of GNAS prompts needs to be aware of the diverse search space and search strategy in GNAS. In particular, the search space in GNAS contains a large number of candidate \textit{operations} and candidate \textit{connections} between operations. The purpose is to generate previously unseen and better architectures from the search space by using LLMs. Figure~\ref{fig:model_fig} gives an overview of the GNAS-LLM method, and we will discuss three important questions in the following. 

\begin{figure}[ht]
      \centering
      \includegraphics[width=0.5\linewidth]{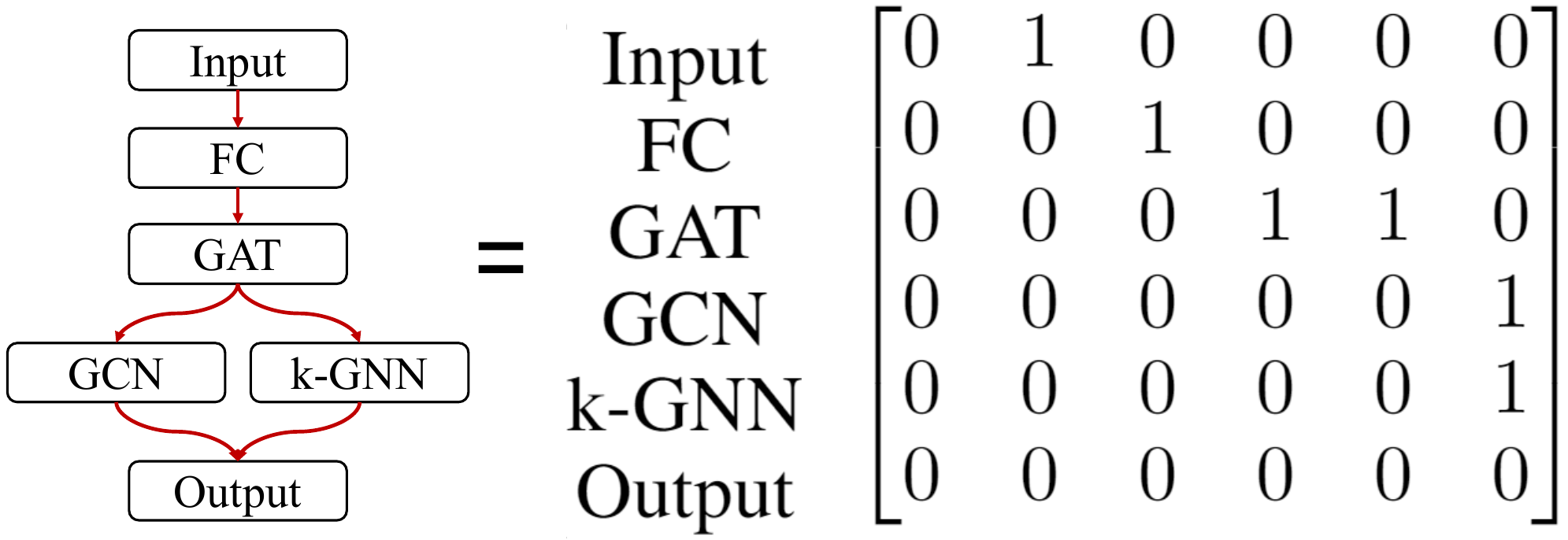}
      \caption{An illustration of a generated architecture (left) represented by an adjacent matrix (right) in which each element ``1" represents a connection between operations.}
      \label{fig:cora_arch}
\end{figure}

\textbf{How can LLMs gain awareness of the search space in GNAS?} The search space in GNAS contains candidate operations and candidate connections between operations. As shown in Figure~\ref{fig:cora_arch}, we define an adjacency matrix to describe connections between candidate operations. Then, the connections are represented by the adjacency matrix and the operations are included into the GNAS prompts, so as to LLMs can understand the search space. 

First, we describe candidate connections in the GNAS prompts. A GNN architecture can be taken as a sample from the search space. The architecture (sample) can be depicted by a Directed Acyclic Graph (DAG), where each node represents an operation and each edge represents a connection. For example, Figure~\ref{fig:cora_arch} shows a GNN architecture consisting of four operations between input and output, utilizing the adjacency matrix to describe the operation connections. The connection pattern remains consistent for all the GNNs within the search space.

Besides candidate connections, we describe candidate operations in the GNAS prompts. Because candidate operations contain irregular message aggregation functions, we directly include these functions in the GNAS prompts and expect LLMs to understand them. Concretely, we include into the GNAS prompts  all the candidate operations obtained  from the NAS-Bench-Graph dataset~\cite{qin2022bench}. Typical operations from NAS-Bench-Graph are GCN, GAT, GraphSAGE, GIN, ChebNet, ARMA,  k-GNN, skip connection, and fully connected layer. 
Taking the operation GAT for example, we include the following functions in the prompt. 
\begin{small}
\begin{mdframed}[linewidth=0.5pt]
$$\displaystyle  \mathbf{x}_i^{\prime}=\alpha_{i, i} \Theta \mathbf{x}_i+\sum_{j \in \mathcal{N}(i)} \alpha_{i, j} \Theta \mathbf{x}_j$$
$$\displaystyle \alpha_{i, j}=\frac{\exp \left(\operatorname{LeakyReLU}\left(\mathbf{a}^{\top} \left[ \boldsymbol{\Theta} \mathbf{x}_i \| \Theta \mathbf{x}_j\right]\right)\right)}{\sum_{k \in \mathcal{N}(i) \cup\{i\}} \exp \left(\operatorname{LeakyReLU}\left(\mathbf{a}^{\top}\left[\Theta \mathbf{x}_i \| \Theta \mathbf{x}_k\right]\right)\right)}$$
\end{mdframed}
\end{small}

\begin{table}[hb]
          \caption{The GNAS prompts for GNAS-LLM.}
        \label{tab:function_template}
          \centering
        \fontsize{8}{1\baselineskip}\selectfont
    \renewcommand{\arraystretch}{0.7}
          \begin{tabular}{|p{1.8cm}|p{13.2cm}|} 
            \toprule
             Prompt Name & Prompt Template
             \\ \midrule
          Space  prompt  & \makecell[l]{ // \textbf{Search Task} \\
          \quad The task is to choose the best GNN  architecture on a given dataset. The architecture will be trained and tested \\ on  [\textcolor{blue!80}{Dataset}], and  the objective is to maximize model accuracy.\\ // \textbf{Search Space}\\
            \quad A GNN architecture is defined as follows: 
            The first operation is input, the last operation is output, and the interm-\\ -ediate operations are candidate operations. 
            The adjacency matrix  of operation connections is as follows:
   [\textcolor{blue!80}{Candidate} \\\ \textcolor{blue!80}{Connections}], where   the (i,j)-th  element in the  adjacency matrix denotes that the output of operation $i$ will be used \\ as the input of operation $j$.There are [\textcolor{blue!80}{Candidate Numbers}] operations that can be selected: [\textcolor{blue!80}{Candidate Operations}].} \\ \midrule
        Strategy prompt &\makecell[l]{
        // \textbf{Search Strategy} \\
        \quad At the beginning, when only a few numbers of evaluated architectures are available, use  \textbf{the exploration strategy} \\ to explore the operations. Randomly select a batch of operations for evaluation.
         When a certain amount of evaluated \\ architectures are available, use \textbf{the exploitation strategy} to find the best operations by sampling the best candidate \\ operations from previously generated candidates. 
        }
            \\ \midrule
            Reward prompt&
            \makecell[l]{
             Model [\textcolor{blue!80}{Architecture}] achieves an accuracy of [\textcolor{blue!80}{Accuracy}].\\
            } \\
            \bottomrule
          \end{tabular}
    \end{table}


\begin{algorithm}[h]
        \caption{Search Process of GNAS-LLM}
        \label{algorithm1}
            \begin{algorithmic}[1]
            \REQUIRE
             A pre-trained  \textbf{LLM}; The search space $\mathcal{M}$; The graph dataset $\mathcal{G}$; The number of iterations $T$; The number of GNNs sampled at each iteration $N$ ;
            \ENSURE
            The best GNN architecture $m^*$ ;
             \item [] // Global model list, accuracy list, and the best model 
            \STATE $M = [~], A = [~],  m^* = \emptyset $  \\
             
            \STATE Generate a GNAS prompt $P_{D}$ with search space $\mathcal{M}$,  candidate operations, and search strategy \\

            \item [] // Input LLM with the setting of GNAS
            \FOR{t=1 \TO T}
               \STATE $M_{t}$ = \textbf{LLM}($P_{D}$, $N$)
               \item [] // Evaluation new GNN architectures 
               \STATE Get the accuracy $A_{t}$ of $M_t$ by evaluating on $\mathcal{G}$\\ 
               \STATE $M = M \cup M_{t}$,  $A = A \cup A_{t}$ 
                \STATE Select the best $m^*$ from  $M$ \textit{with respect to} $A$\\

               \item [] // Add reward prompt
               \STATE Generate a reward prompt $P_F$ \textit{with respect to} $M$ and $A$
               \STATE $P_{D}$ = $P_{D} \cup P_F$ 
            \ENDFOR
            \STATE \textbf{return} $m^*$
        \end{algorithmic}
    \end{algorithm}

\textbf{How to guide LLMs to explore the search space of GNAS?} We include into GNAS prompts the description of the search strategy. The prompts guide LLMs to understand  reinforcement learning which consists of exploration and exploitation. In exploration, we instruct LLMs to globally explore the entire space. In exploitation, we instruct LLMs to locally sample the best candidate operations from previously generated candidates. 

Table~\ref{tab:function_template} shows that the GNAS prompts consist of three components, i.e., search task, search space, and search strategy. According to our experiments, these prompts can effectively guide LLMs to find the best graph neural architectures on a given graph dataset.

\textbf{Why GNAS-LLM works?}
The effectiveness of GNAS-LLM originates from several factors. 
First, LLMs are capable of analyzing  graph data, as pointed out by previous work~\cite{zhang2023llm4dyg,huang2023can,wang2024can}. GNAS-LLM, based on LLMs, is also capable of generating  GNN architectures.
Second, LLMs can act as the controller of the GNAS which performs better than existing GNAS controllers. Third, GNAS prompts, by describing the details of the search space and search strategies, enable LLMs to effectively navigate the search space and iteratively generate better GNNs.

Note that GNAS-LLM introduces practical improvements to support the unique challenges of GNN architecture search, comparing GENIUS\cite{zheng2023can}. First, it accommodates the more complex search spaces of GNNs compared to CNNs. GNN operations range from meta-operators in methods like AutoGEL to standard graph convolutional layers, such as those in NASBenchGraph. GNAS-LLM's flexible design ensures compatibility with diverse operations, making it suitable for different levels of complexity. Second, the inclusion of candidate connections in the prompt design allows GNAS-LLM to adapt to complex GNN architectures and different types of search spaces by explicitly representing connections between operations. Third, GNAS-LLM supports flexible search strategies by enabling modifications to the search strategy prompt, making it adaptable to various search methods. These features make GNAS-LLM a versatile tool for GNN architecture search, efficiently addressing operational diversity and structural complexity.

\subsection{Algorithm}
Figure~\ref{fig:model_fig} shows that LLMs can be taken as the controller of GNAS and thus generate new GNN architectures from the search space based on the GNAS prompts. The overall process is outlined in Algorithm \ref{algorithm1}. To initiate GNAS-LLM, we generate a GNAS prompt $P_{D}$ and call the LLM to generate a set of new GNN architectures $M_t$. Based on the output of the LLM, we evaluate the performance of the generated GNNs $M_t$. The evaluation result $A_t$  is taken as the reward, guiding the LLM  towards generating better GNN architectures in the subsequent iterations. By continuously integrating the generated GNNs $M$ and their evaluation results $A$ in the reward prompt $P_F$, the LLM converges fast. In the last step, GNAS-LLM obtains the best GNN $m^{*}$ generated by the LLM.

\section{Experiments}
\label{sec:Exp}

In this section, we conduct experiments to validate the performance of GNAS-LLM. First, we evaluate the results on a search space with different candidate connections. Second, we test the performance with respect to different GNAS prompts by changing the candidate operations. Third, we compare GNAS-LLM with existing GNAS methods. 
Furthermore, we conduct case studies to show the utility of our approach under complicated scenarios, including comparisons with differentiable NAS within the AutoGEL search space,  node classification on homogeneous graphs, and link prediction on heterogeneous graphs. We also evaluate GNAS-LLM across various LLM configurations and use the search space of Pasca on large graphs.


\subsection{Experiment Setup}


\subsubsection{NAS-Bench-Graph Benchmark}
We build the GNAS search space based on the NAS-Bench-Graph Benchmark~\cite{qin2022bench}. There are a total number of nine operation connections in the NAS-Bench-Graph benchmark. Also, the benchmark contains a large number of generated graph architectures, which allows the performance of the generated GNN architectures to be directly queried without actually training the GNN models. The NAS-Bench-Graph benchmark provides accuracy, rank, parameter sizes, and running time of the generated GNNs on datasets such as Cora, Citeseer, Pubmed, and ogbn-arXiv.

\subsubsection{Baselines}
We compare GNNs designed by GNAS-LLM with all the seven  GNNs listed in the NAS-Bench-Graph benchmark. The best results of those GNNs are taken as the baselines. We also compare the proposed model with other GNAS methods, including  Random Search, GraphNAS~\cite{DBLP:conf/ijcai/GaoYZ0H20},   and Genetic-GNN~\cite{Shi2020EvolutionaryAS}. 
    
\subsubsection{Hyperparameters}
GNAS-LLM runs 15 search iterations. The number of iterations is constrained by the length of prompts  that the LLMs allow. During each iteration, LLMs generate 10 new architectures. For each architecture search, we only use one candidate operation connections and nine candidate operations.
In the following experiments, we repeat our method three times and show the best results \textit{w.r.t.} accuracy on a validation dataset.

If not otherwise specified, we use GPT-4 with version V20230314 as the default LLM in the experiments. For all models, we set temperature $\tau = 0$ for reproducibility. We adopt accuracy as the metric for all tasks. Additionally, we use all the GNAS baselines to generate $N=10$  architectures at each iteration. Specifically, for Genetic-GNN, the initial population is set to 50, and the number of parent individuals selected at each iteration is 15. For GraphNAS, the ADAM optimizer is used, with a learning rate of 0.00035.
In the experiments involving AutoGEL for the dataset of Cora, Citeseer, and Pubmed, we set the layer number to 2, the ADAM optimizer with a learning rate of 5e-4, a minibatch size of 128, and train each generated architecture for 200 epochs with a dropout value of 0.5. And for the dataset of FB15k-237~\cite{toutanova2015observed} and WN18RR~\cite{dettmers2018convolutional}, we set 1 layer, the use of the ADAM optimizer with a learning rate set at 0.001, and a minibatch size of 128. We trained each generated architecture for a total of 200 epochs, implementing a dropout value of 0.1.




\subsection{Results on Search Space }


\begin{table}[htb]
    \centering
    \caption{Results of GNAS-LLM on nine search spaces in NAS-Bench-Graph \textit{w.r.t.} accuracy (\%) and the corresponding rank  within the search space (numbers in parentheses indicate the rank). The best results are in bold.}
    \label{tab:experiment2}
    \setlength{\tabcolsep}{5pt}
    \fontsize{8}{1\baselineskip}\selectfont
    \renewcommand{\arraystretch}{0.7}
    \begin{tabular}{cccccccccc}\toprule
        Method & Space 1 & Space 2 & Space 3 & Space 4 & Space 5 & Space 6 & Space 7 & Space 8 & Space 9 \\ \midrule
        Random & \textbf{81.00 (1)} & 81.37 (64) & 81.47 (20) & \textbf{82.37 (1)} & 81.40 (42) & 81.27 (17) & 81.97 (6) & 81.07 (30) & 81.03 (9) \\
        \fontsize{8}{1\baselineskip}\selectfont GraphNAS & 80.80 (2) & 81.77 (22) & \textbf{81.80 (5)} & 81.53 (20) & 81.57 (28) & 81.67 (4) & 81.43 (44) & 81.80 (3) & 81.03 (9) \\
        \fontsize{8}{1\baselineskip}\selectfont Genetic-GNN & 80.63 (5) & 82.27 (3) & \textbf{81.80 (5)} & 81.73 (12) & 82.00 (10) & \textbf{81.87 (2)}& \textbf{82.37 (1)}& 81.43 (12) & 81.47 (4) \\
        \fontsize{7}{1\baselineskip}\selectfont GNAS-LLM & \textbf{81.00 (1)} & \textbf{82.37 (2)} & \textbf{81.80 (5)} & \textbf{82.37 (1)} & \textbf{82.37 (5)} & \textbf{81.87 (2)} & \textbf{82.37 (1)} & \textbf{83.13 (1)} & \textbf{81.70 (2)} \\
        \bottomrule
    \end{tabular}
\end{table}

\begin{figure*}[ht]
      \centering
      \includegraphics[width=1\linewidth]{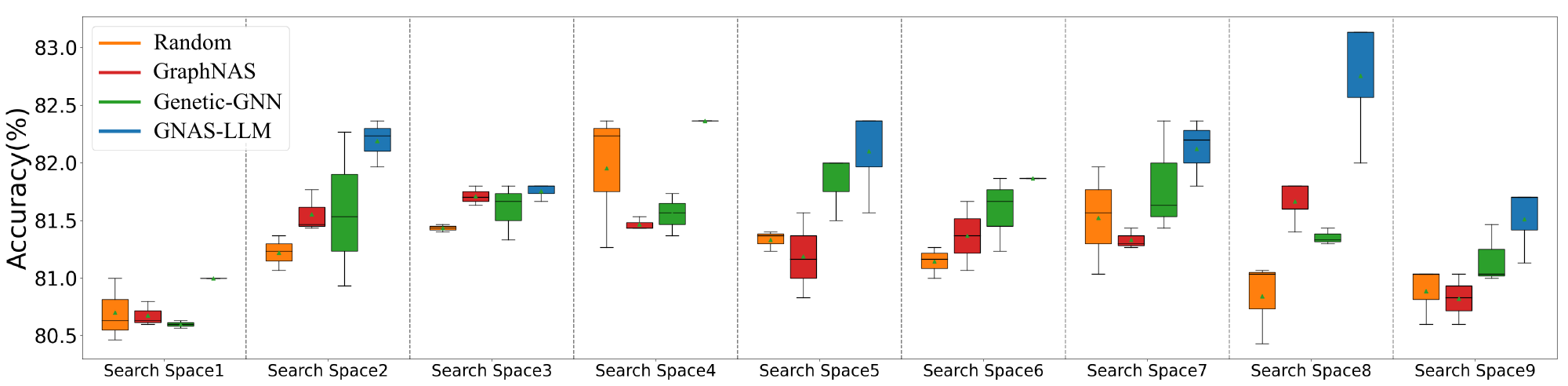}
      \caption{The accuracy (\%) of the GNNs  designed by GNAS-LLM and  baselines on nine search spaces of Cora. Comparing with other baselines,  the GNNs designed by our method achieve the highest scores in all search spaces \textit{w.r.t.} the average accuracy.}
      \label{fig:ablation_searchspace}
    \end{figure*}

    To validate the performance of GNAS-LLM, we run the architecture search algorithm three times using all of the nine candidate operation connections from the NAS-Bench-Graph search spaces.   We then compare our method with other GNAS methods. 
    Table~\ref{tab:experiment2} shows the accuracy and ranks of the designed GNNs. Given a search space in NAS-Bench-Graph, a rank refers to the accuracy ranking of the designed GNNs within that space. Figure~\ref{fig:ablation_searchspace} shows the mean accuracy and variance of the designed GNNs.


As shown in Table~\ref{tab:experiment2}, our approach achieves an average accuracy of 82.11\% across 9 search spaces, and surpassing the other three baseline methods, where Random achieves  better results than the other two baselines in Spaces 1 and 4, GraphNAS ranks top in Space 3,  and Genetic-GNN ranks top in Spaces 3, 6, and 7. Note that our method is competitive to the best baselines on all the nine spaces. In particular,  GNAS-LLM wins all the other baselines  by 0.51\% on average in Spaces 2, 5, 8, and 9. The most successful result is on Space 8, where the result beats the best baseline GraphNAS by improving the accuracy of 1.33\%.

The results demonstrate that GNAS-LLM achieves top-tier results across all the nine different search spaces. These results emphasize that GNAS-LLM performs well on different operation connections and consistently yields satisfactory models. Figure~\ref{fig:ablation_searchspace} shows that our method achieves higher average accuracy results than the other methods under the nine different search spaces. In particular, our method finds  the best GNN models in four search spaces  (e.g., Spaces 1, 4, 7, and 8). The results show that our method is capable of designing the best model for a new search space with different operations connections.

\subsection{Results on GNAS prompts}



    
    To validate the performance of the GNAS prompts, we construct two types of variants of GNAS-LLM for ablation study. 

    First, we test variants by removing parts of the GNAS prompt. We design three variants, namely '$\neg Connections$', '$\neg Operations$', and '$\neg Strategy$', to denote prompts without the corresponding description.  '$\neg Operations$' denotes prompts without the description of the operations.   Table~\ref{tab:experiment4} shows the results of GNAS-LLM and the variants. 
    Obviously, when excluding any item of the search space and search strategy descriptions, the variants can still output satisfactory results. However, these variants output worse results than GNAS-LLM. 
    On four datasets, GNAS-LLM exhibits a modest average accuracy improvement of 0.52\% over the variants, alongside a ranking advancement of 15 places. The variant $\neg Operations$ consistently outperforms or, at the very least, equals the other variants across all four datasets. Conversely, the $\neg Connections$ variant lags behind other variants on all datasets, which may suggest the relative importance of different components to a certain extent.
    According to the above results, we can conclude that all the parts of the GNAS prompts (e.g., space prompt and strategy prompt) are helpful in generating GNN architectures, which suggests that the GNAS prompts are capable of avoiding local optima by iteratively run the GNAS prompts. 

    Second, we test two variants of the GNAS prompts by using new search strategies and candidate connections.
    We construct the `with $Evolutionary$' variant by replacing the search strategy of the GNAS prompt with evolutionary algorithms. 
    Furthermore, we construct the `with $Tuple$' variant by describing the candidate operation of GNN architecture using the operation tuple list.
    Note that the design of the `with $Evolutionary$' variant aims to enable LLMs to autonomously design candidate operations by organizing the edges within the operation tuple list.
    As shown in Table~\ref{tab:experiment4}, the variant `with $Evolutionary$'  achieves comparable results with respect to GNAS-LLM on the Citeseer dataset, while performing below GNAS-LLM by an average of 0.58\% on the remaining three datasets. This shows that the original search strategy is more adept at handling these three datasets than the evolutionary search strategy. 
    Meanwhile, the variant `with $Tuple$' underperforms GNAS-LLM on all four datasets, with an average drop of 0.75\%. This indicates that LLMs are capable of understanding the description of a model architecture using the adjacency matrix and the operation lists.

    \begin{table}[ht]
    \centering
    \caption{Results of the variants of GNAS prompts \textit{w.r.t.} accuracy (\%) and the corresponding rank within the search space
(numbers in parentheses indicate the rank). The best results are in bold.}
    \label{tab:experiment4}
    \fontsize{8}{1\baselineskip}\selectfont
    \renewcommand{\arraystretch}{0.7}
    \begin{tabular}{ccccccc}
        \toprule
         Dataset & GNAS-LLM & $\neg Operation$ & $\neg Connections$ & $\neg Strategy$ & with $ Evolutionary$ & with $ Tuple$ \\ 
        \midrule
        Cora & \textbf{83.13 (1)} & 82.00 (26) & 81.80 (49) & 81.80 (49) & 82.00 (26) & 81.47 (171) \\ 
        Citeseer & \textbf{71.37 (2)} & 70.80 (20) & 70.17 (119) & 70.80 (20) & \textbf{71.37 (2)} & 70.67 (30) \\ 
        Pubmed & \textbf{78.30 (3)} & 78.03 (11) & 78.03 (11) & 77.90 (16) & 78.00 (12) & 77.80 (33) \\ 
        arXiv & \textbf{72.39 (1)} & 72.28 (9) & 72.07 (98) & 72.28 (9) & 72.08 (98) & 72.27 (10) \\ 
        \bottomrule
    \end{tabular}
\end{table}

    Thirdly, to investigate whether adding the name of the dataset in our prompt would lead to information leakage, we construct a variant of `$\neg Dataset$' by removing all related information about the dataset from our prompt. The results of GNAS-LLM and its variants show competitive results. 
    They achieve the same results on the Cora and Citeseer datasets, with accuracy of 83.13\% and 71.37\%, respectively. On the ArXiv dataset, GNAS-LLM outperforms `$\neg Dataset$' by a margin of 0.12\% in terms of accuracy. However, it lags by 0.30\% in accuracy on the Pubmed. From the results,  we can conclude that revealing the name of the dataset in the prompt does not significantly cause information leakage.

    \begin{figure*}[!ht]
      \centering
       \includegraphics[width=0.99\linewidth]{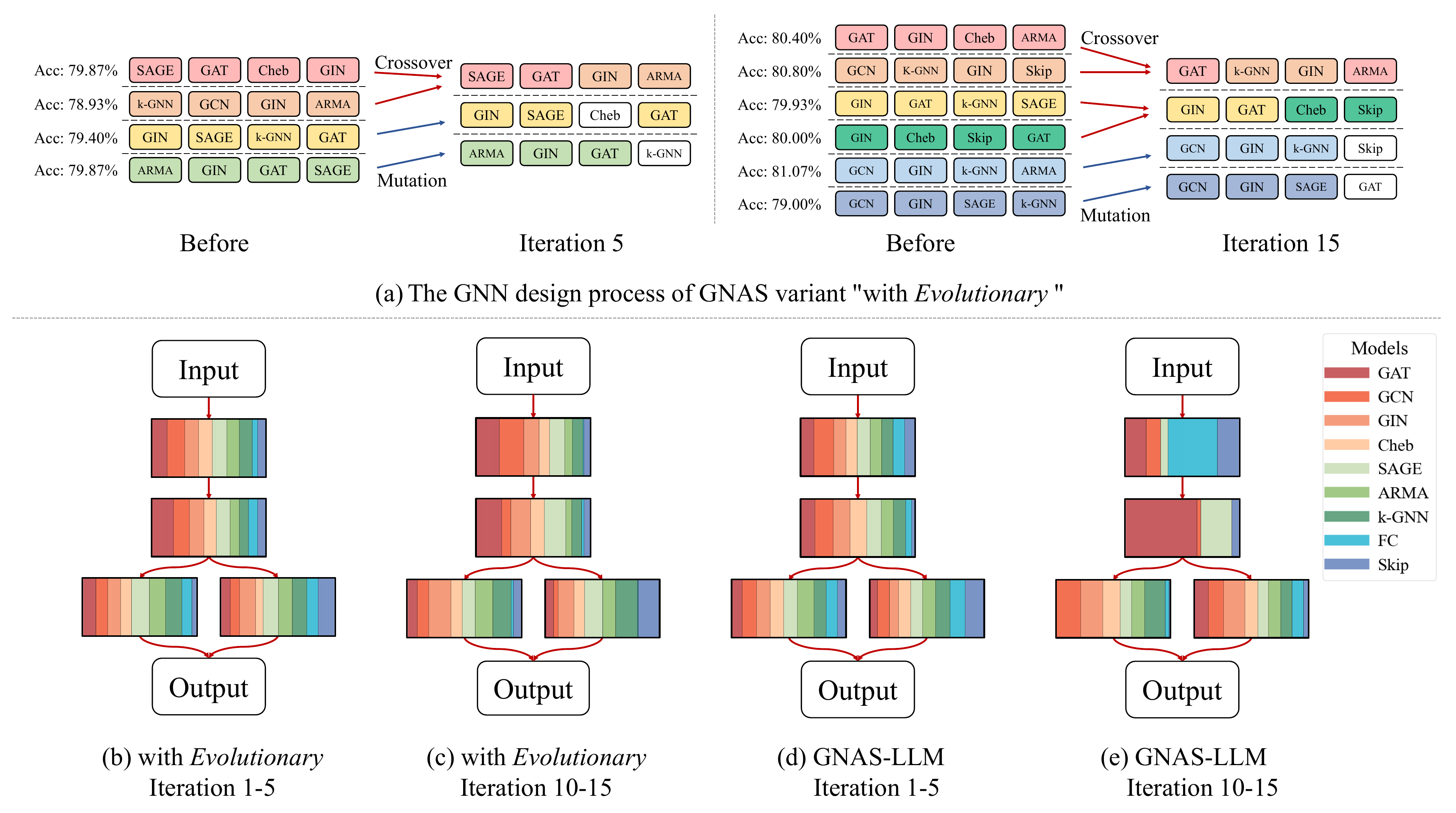}
      \caption{Analysis of search strategy prompts in GNAS-LLM. \textbf{(a)} LLMs can comprehend and execute the given search strategy prompt. The `with $Evolutionary$' variant follows the evolutionary search strategy prompt and generates new architectures by applying crossover and mutation to high-performing GNN architectures. \textbf{(b)-(e)} The percentage of GNN operations selected by LLMs under different search strategy prompts. The results highlight the influence of varying search strategies on LLMs during architecture search.}
      \label{fig:rl_evo}
    \end{figure*}
    
    Finally, to evaluate the ability of LLMs to understand and follow specified search strategies, we conducted experiments analyzing the impact of search strategy prompts, as shown in Figure~\ref{fig:rl_evo}. Initially, we employed the evolutionary search strategy prompt as a case study to assess whether LLMs can faithfully execute the principles of an evolutionary algorithm. Figure~\ref{fig:rl_evo}(a) demonstrates that the GNAS-LLM variant with `$Evolutionary$', utilizing the evolutionary search strategy prompt, successfully adheres to the evolutionary search strategy throughout the GNAS process.  Furthermore, to investigate the influence of different search strategy prompts, we analyzed the distribution of GNN operations designed by the LLM under varying strategies, as illustrated in Figures~\ref{fig:rl_evo}(b), (c), (d), and (e). Specifically, in Figure \ref{fig:rl_evo}(b), the with `$Evolutionary$' variant is prompted to maximize exploration of diverse GNN architectures during the initial stage; in Figure \ref{fig:rl_evo}(c), it adheres to the evolutionary algorithm search strategy; in Figure \ref{fig:rl_evo}(d), it prioritizes an exploration  search strategy; and in Figure \ref{fig:rl_evo}(e), it emphasizes the exploitation strategy. The experimental results  demonstrate that the behavior of LLMs in architecture search is  modulated by the choice of search strategy prompt.


\subsection{Results on Graph Neural Architecture Search}

\begin{table}[h]
  \caption{Results of Graph Neural Architecture Search on NAS-Bench-Graph \textit{w.r.t.} accuracy(\%) and the corresponding rank  within the search space (numbers in parentheses indicate the rank). The best results are in bold. The second best results are underlined.}
   \label{tab:labelexperiment1}
  \centering
      \fontsize{8}{1\baselineskip}\selectfont
    \renewcommand{\arraystretch}{0.7}
    \begin{tabular}{ccccccccc}
    \toprule
     \multirow{2}{*}{Method}  & \multicolumn{2}{c}{Cora} & \multicolumn{2}{c}{Citeseer} &\multicolumn{2}{c}{Pubmed} & \multicolumn{2}{c}{arXiv} 
     \\\cmidrule(lr){2-3}\cmidrule(lr){4-5}\cmidrule(lr){6-7}\cmidrule(lr){8-9}
        & Val & Test & Val & Test & Val & Test & Val & Test   \\
         \midrule
         ChebNet & 79.33 & 77.33 & 67.30 & 69.00 & 75.37 & 75.13 & \underline{72.31} & \textbf{73.53}      \\ \midrule
        GCN        & 82.27 & 79.00  & 69.10 & \textbf{71.13} & 77.47 & 78.13 & 72.03 & 73.10   \\ 
        GraphSAGE  & 80.13 & 78.47 & 68.80 & 69.53 & 77.13 & 77.93 & 71.97 & 72.70 \\   
        GAT        & 81.80 & 79.73 & 69.27 & 68.73 & 77.20 & 78.67 & 71.20 & 73.10 \\ 
        GIN        & 79.83 & 78.93 & 68.50 & 68.33 & 75.93 & \underline{79.67} & 66.60 & 67.80 \\ 
        k-GNN    & 78.40 & 77.07 & 66.60 & 65.73 & 74.50 & 78.33 & 67.95 & 69.07     \\ 
        ARMA       & 79.17 & 76.27 & 66.03 & 68.67 & 75.50 & 75.53 & 71.76 & 73.07 
         \\\midrule
        Random-NAS & \underline{82.37 (8)} & \underline{79.80 (3139)} & \underline{70.66 (29)} & 70.13 (1449) & 77.63 (57) & \textbf{80.20 (55)} & 72.18 (32)  & 73.33 (60) \\
        GraphNAS & 81.80 (49)  & 79.60 (3997)  & 70.56 (35) & 70.20 (1222) & \underline{78.27 (4)} & 78.47 (5071) & 72.10 (79) & \underline{73.50 (9)}  \\
        Genetic-GNN& \underline{82.37 (8)} & \underline{79.80 (3139)} & 70.67 (30) & \underline{70.93 (261)} & \underline{78.27 (4)} & 78.47 (5071) & 72.21 (18) & \textbf{73.53 (5)} \\
        \textbf{GNAS-LLM}  & \textbf{83.13 (1)} & \textbf{80.93 (84)} & \textbf{71.37 (2)} & 70.07 (1546) & \textbf{78.30 (3)} & 79.33 (1031) & \textbf{72.39 (1)} & 73.33 (61) \\
         \bottomrule
  \end{tabular}
\end{table}    

To evaluate the performance of GNAS-LLM on graph neural architecture search, we compare GNAS-LLM with reinforcement learning  GNAS, evolutionary GNAS, and differential gradient GNAS on Cora, Citeseer, Pubmed, and arXiv.  We compare the GNNs discovered by our method with these baseline GNAS methods. 

Table~\ref{tab:labelexperiment1} shows the comparisons between the GNNs discovered by GNAS-LLM and other GNAS method \textit{w.r.t.}  the accuracy and rank of GNNs on NAS-Bench-Graph. Obviously, the GNNs models generated by our method are  at the top of the entire search space in terms of validation accuracy.  
The results confirm the efficacy of our method in designing GNNs. Specifically, our approach achieves an average of 0.42\% accuracy improvement on the validation dataset compared with the baseline methods. In particular, our method can design the best GNN architectures for both Cora and arXiv \textit{w.r.t.} accuracy on the validation set. 

While the generated architectures show exceptional results on the validation set, their results on the test set may not be always the best. Furthermore, considering the Pubmed dataset,  the best model on the validation set ranks 3, whereas on the test set drops to 1,031. 
This suggests a obvious gap between the validation and test sets, highlighting the need for a more precise and robust model evaluation approach to GNAS. Utilizing methods such as variance analysis on the validation set and incorporating confidence intervals for selection purposes could enhance the accuracy and reliability of model evaluations in the GNAS tasks.

\begin{table}[h]
  \caption{Results of three architecture search methods \textit{w.r.t.} Accuracy(\%). The best results are in bold. The second best results are underlined.}
   \label{tab:average3}
  \centering
    \fontsize{8}{1\baselineskip}\selectfont
    \renewcommand{\arraystretch}{0.7}
  \begin{tabular}{ccccccccc}
    \toprule
      \multirow{2}{*}{Method} & \multicolumn{2}{c}{Cora}& \multicolumn{2}{c}{Citeseer}&\multicolumn{2}{c}{Pubmed}&\multicolumn{2}{c}{arXiv}\\ 
      \cmidrule(lr){2-3}\cmidrule(lr){4-5}\cmidrule(lr){6-7}\cmidrule(lr){8-9}
      & Val & Test& Val & Test& Val & Test& Val & Test\\ \midrule
Random      & 80.98±0.13&79.69±0.31&70.16±0.44&69.96±0.37&77.43±0.24& \underline{79.49±0.91} & 72.14±0.07 & 73.31±0.04  \\ 
GraphNAS    & 81.01±0.10 & 79.69±0.20& 70.01±0.48 & \underline{70.11±0.21}& 77.80±0.52 & 79.36±0.84  & 72.03±0.06 & 73.33±0.17  \\ 
Genetic-GNN& \underline{81.17±0.54} & \underline{80.27±0.68}& \underline{70.28±0.40} & 70.02±1.00& \underline{77.90±0.33 }& 78.89±1.00& \underline{72.15±0.08} & \textbf{73.41±0.13 }   \\ 
\textbf{GNAS-LLM}& \textbf{82.76±0.65} &\textbf{80.93±0.00}& \textbf{71.19±0.31} &\textbf{70.22±0.27}& \textbf{78.03±0.23} &\textbf{79.87±0.46} & \textbf{72.29±0.09} &\textbf{73.41±0.07}  \\
         \bottomrule
  \end{tabular}
\end{table} 

In Table~\ref{tab:average3}, we show  comparison of the performance of the architecture search under different validation and test sets. Our method procures favorable results on both test and validation sets across four datasets, with an average increase of 0.74\% on the validation sets and of 0.29\% on the test sets.

\begin{figure*}[t]
      \centering
      \includegraphics[width=1\linewidth]{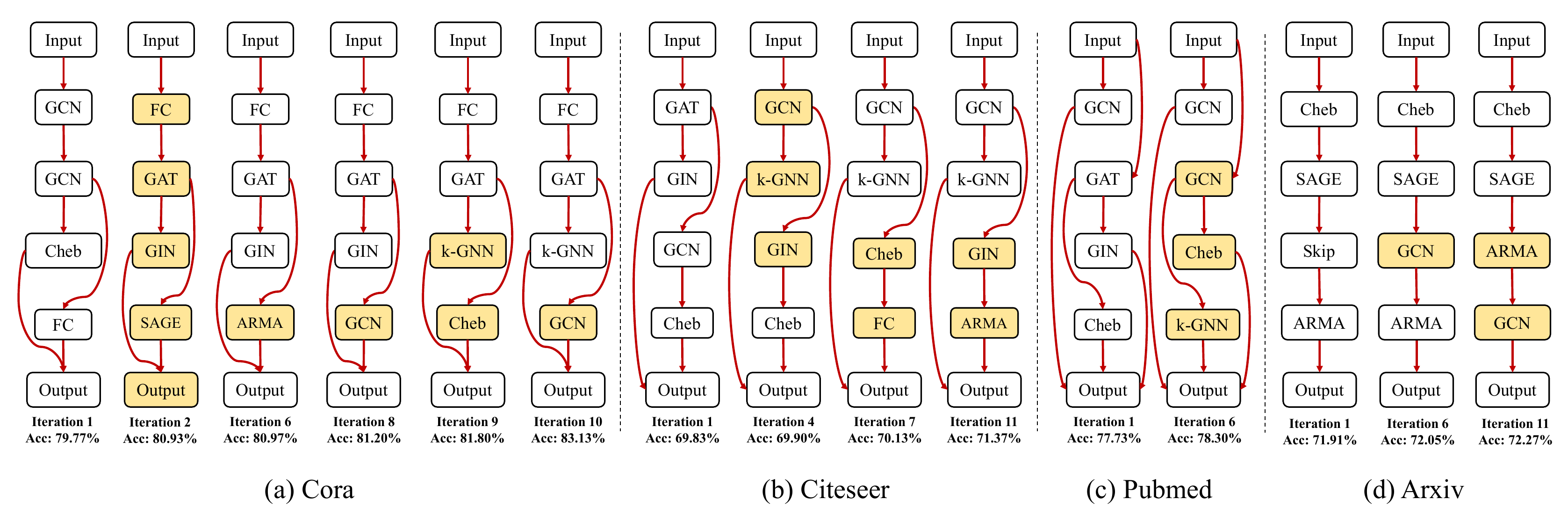}
      \caption{The GNNs generated by GNAS-LLM during search iterations. The colored blocks indicate changes to the previous model.}
      \label{fig:arch_development}
\end{figure*}

Moreover, we test the dynamic evolution of the best architecture on the datasets. The results are shown in Figure~\ref{fig:arch_development}. The GNAS-LLM refines the designed GNN architecture. Once the backbone architecture is set, the operations from the existing best architecture continuously replace, adapt, and refine the architecture. Taking the experiments on Cora as an example, GNAS-LLM initially randomly selects several architectures for performance evaluation, from which it identifies architectures with better performance, such as the first and second architectures in Figure~\ref{fig:arch_development}a. Subsequently, for the existing higher-performing architectures, some operations are fixed, while others are replaced for exploration, as exhibited in the 3-rd,4-th,5-th,6-th architecture in Figure~\ref{fig:arch_development}a. GNAS-LLM fixes the first two operations in this architecture and replaces the final two operations,  deriving a better architecture.

\begin{figure}[h]
  \centering
    \includegraphics[width=1\linewidth]{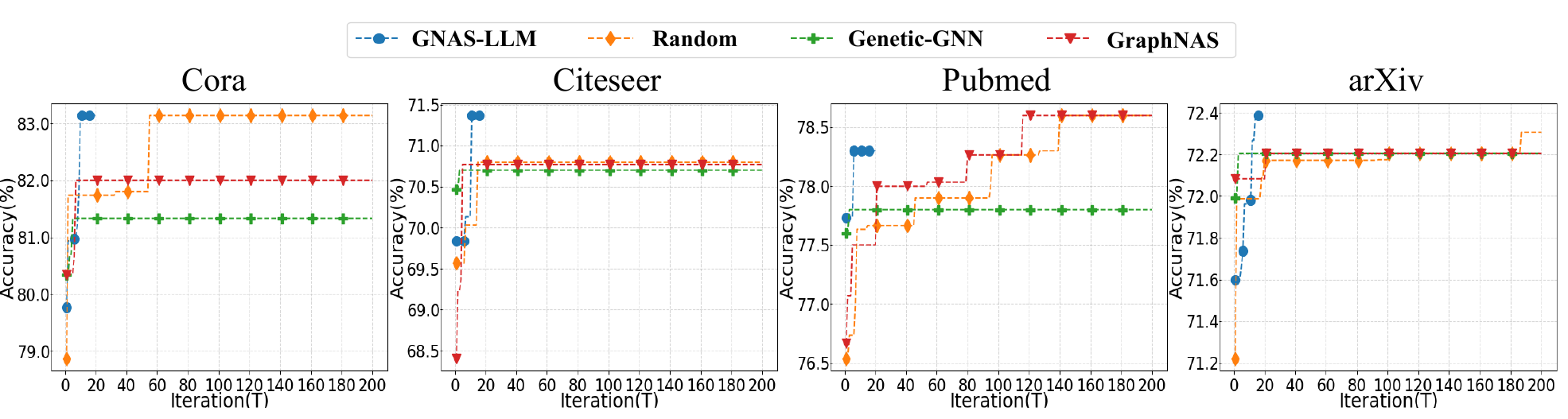}
  \caption{Results of the baseline methods with 200 architecture search iterations. GNAS-LLM runs less than 15 search iterations and then converges to the best GNN architectures. }
  \label{fig:200times}
\end{figure}

\begin{figure}[h]
  \centering
    \includegraphics[width=1\linewidth]{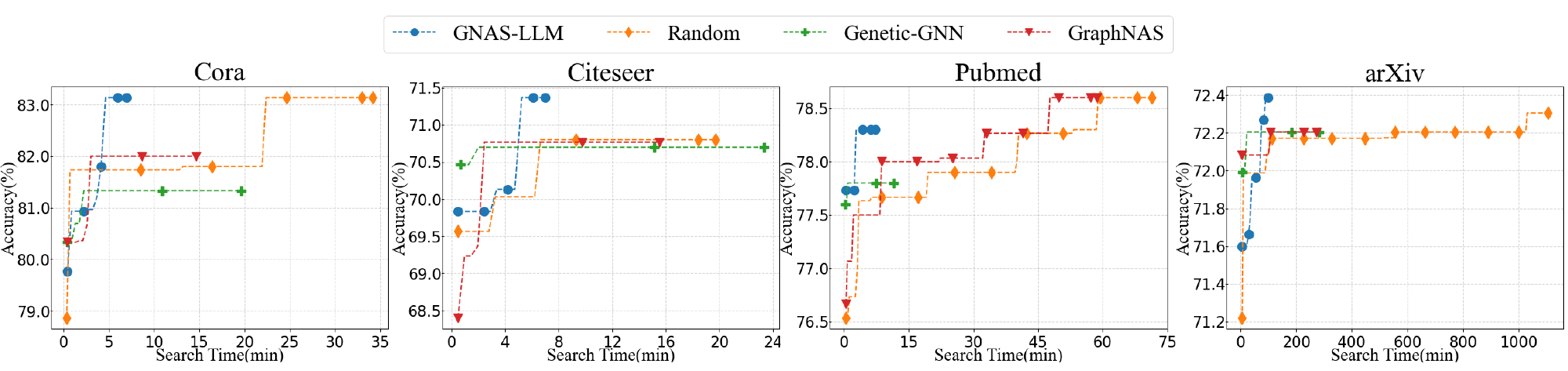}
  \caption{Results of GNAS-LLM and the baseline methods with respect to search time cost.}
  \label{fig:acc_time}
\end{figure}

In order to showcase the advantage of our method at the convergence, we set the search iterations of the baselines to 200,  while our approach only carries out 15 iterations of the architecture search, because the input tokens are limited by GPT-4. The results are shown in Figure~\ref{fig:200times}. The experiments on  Citeseer and arXiv show that our method surpasses all the baselines because the baseline method cannot find a better GNN than our method even with increasing search iterations. Moreover, experiments on Cora show that the Random method can find the  GNN comparable to our method by using more than 50 iterations.  Additionally, experiments on Pubmed show that Random and GraphNAS can find GNNs with more than 80 iterations and achieve 0.2\% improvement in accuracy compared to our method. In conclusion, our method can find well-performance GNNs   within 15 iterations on the benchmark datasets.  In other words, our approach will explore fewer candidate GNNs to find the best GNN architectures.

To further showcase the time efficiency of our approach, we conducted a comparison with respect to the architecture validation time with the baseline methods of 200 iterations. Our approach requires only 15 architecture search iterations.   As depicted in Figure ~\ref {fig:acc_time}, our method converges to the final results faster than all the baselines, including the communication time with the GPT-4 server. The communication time with the server is almost negligible, particularly for large datasets.

We compare the top 10 candidate GNNs designed by our method and the baselines during the architecture search. 
Experimental results show that the GNNs designed by our method are closest to the best GNNs on the validation set. For example, on Cora and arxiv, the top-two architectures generated by our method are better than the baselines \textit{w.r.t.} accuracy on the validation set. Moreover, on Cora, Pubmed, and arxiv, one of the GNNs generated by our method achieves the best accuracy result on the test set, compared with the GNNs designed by baselines.

\subsection{Case study 1: Comparing with  AutoGEL}

We expand GNAS-LLM into the search space within AutoGEL to demonstrate the superiority of our approach. Because Autogel runs experiments on Cora, Citeseer, and Pubmed, we conduct comparisons also on these datasets, using the search space and experimental configurations of AutoGEL. The results of the AutoGEL method are obtained by executing their codes under our testing environment.

\begin{table*}[h!]
\centering
    \caption{Results of architecture search with the AutoGEL search space. The best results are in bold.}
    \label{tab:AUTOGEL}
    \setlength{\tabcolsep}{4pt}
    \fontsize{8}{1\baselineskip}\selectfont
    \renewcommand{\arraystretch}{0.7}
    \begin{tabular}{lccccccccc}
        \toprule
        \multirow{2}{*}{Dataset} &\multirow{2}{*}{Method}& \multicolumn{2}{c}{Top 1} & \multicolumn{2}{c}{Avg. of Top 2} &\multicolumn{2}{c}{ Avg. of Top 5} & \multicolumn{2}{c}{Avg. of Top 10} 
         \\\cmidrule(lr){3-4}\cmidrule(lr){5-6}\cmidrule(lr){7-8}\cmidrule(lr){9-10}
                &  & Val & Test & Val & Test & Val & Test & Val & Test   \\
        \midrule 
    \multirow{2}{*}{Cora}&AutoGEL&\textbf{89.48±0.00}&89.30±0.00&89.24±0.25&89.58±0.28&88.98±0.30&89.56±0.41&88.55±0.51&89.70±0.40\\ \cmidrule(lr){2-10} 
    &GNAS-LLM&\textbf{89.48±0.00}&\textbf{91.51±0.00}&\textbf{89.48±0.00}&\textbf{91.14±0.37}&\textbf{89.38±0.09}&\textbf{90.73±0.42}&\textbf{89.34±0.09}&\textbf{90.62±0.42}\\ \midrule 
    \multirow{2}{*}{Citeseer} &AutoGEL&74.59±0.00&77.33±0.12&74.59±0.00&77.60±0.29&74.39±0.18&\textbf{77.84±0.30}&73.97±0.48&77.43±0.60\\ \cmidrule(lr){2-10} 
    &GNAS-LLM&\textbf{75.19±0.00}&\textbf{78.08±0.00}&\textbf{75.11±0.08}&\textbf{77.70±0.38}&\textbf{75.08±0.10}&77.33±0.39&\textbf{74.93±0.18}&\textbf{77.55±0.55}\\ \midrule 
    \multirow{2}{*}{Pubmed} &AutoGEL&89.19±0.05&89.53±0.02&89.18±0.11&89.57±0.06&89.05±0.18&89.55±0.10&88.82±0.29&89.48±0.19\\ \cmidrule(lr){2-10} 
    &GNAS-LLM&\textbf{89.48±0.09}&\textbf{89.62±0.04}&\textbf{89.45±0.09}&\textbf{89.64±0.09}&\textbf{89.39±0.08}&\textbf{89.66±0.17}&\textbf{89.34±0.10}&\textbf{89.67±0.20}\\
        \bottomrule
    \end{tabular}
\end{table*}

The results in Table~\ref{tab:AUTOGEL} indicate that the best GNN architecture  designed by our approach outperforms AutoGEL on the validation sets and test sets on all the three datasets. 
Our approach shows average performance enhancement of 0.3\% on the Validation set and 1.02\% on the Test set in comparison to the best-designed GNNs. The average improvement for the top two models are 0.34\% on Validation and 0.58\% on Test, while the top five models exhibit average increase of 0.48\% on Validation and 0.59\% on Test. Moreover, the top ten models demonstrate an average improvement of 0.76\% on Validation and 0.41\% on Test.
The best performance improvement is from the Cora dataset, where the improvement of the best model reaches 2.21\%. On the test set of Citeseer, the results of the top five models are inferior to the baseline. 

To sum up, the case study on AutoGEL search space shows that our method can efficiently adapt to new search spaces and design the best GNNs. Furthermore, our method outperforms the differential-based GNAS method such as AutoGEL in terms of accuracy.

\subsection{Case study 2: Comparing with other LLMs}
    We conduct experiments on multiple LLMs, such as ChatGLM3-6B\cite{zeng2022glm}, GPT-3.5-turbo~\cite{brown2020language},  and PaLM 2~\cite{anil2023palm}. For the GPT-3.5-turbo and Palm 2 models, we call their online APIs, while for the ChatGLM3-6B model, we download and run locally.

\begin{figure}[t]
  \centering
  \centering
    \includegraphics[width=0.9\linewidth]{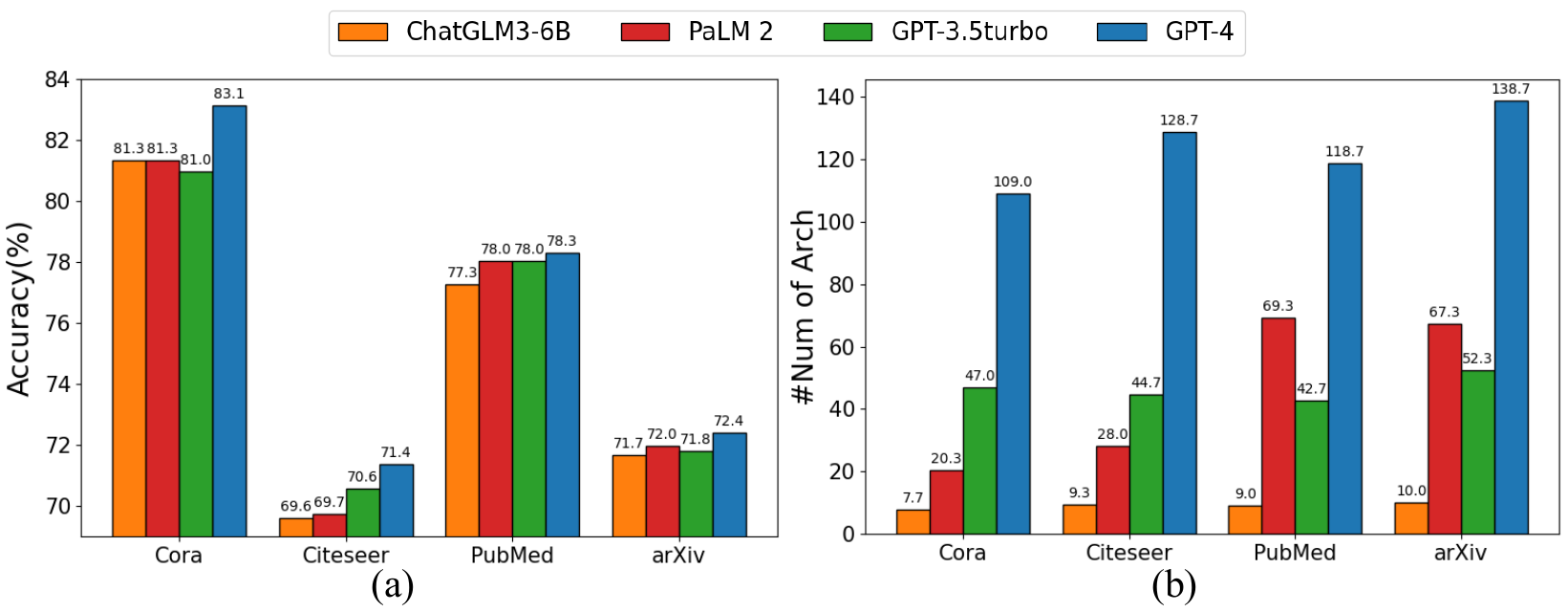}  
  \caption{Results of case study on architecture searching with different LLMs. (a)The accuracy of GNNs designed by different LLMs. (b) The number of distinct architectures generated by LLMs during architecture search.}
  \label{fig:iso_perf}
\end{figure}

    The results of the experiments are shown in Figure~\ref{fig:iso_perf}a. GPT-4  achieves the best performance across all the datasets. On average, GPT-4 obtains a raise of 0.83\% compared to other models, and ranks higher on the four datasets. The best result is reported on the Cora dataset, with a raise of 1.8\%.
    On Pubmed,  PaLM 2 is slightly better than GPT-3.5-turbo. PaLM 2 performs better on the arXiv and Cora datasets. Conversely, GPT-3.5-turbo outperforms PaLM 2 on the Citeseer dataset. The ChatGLM model has the worst results on all three datasets except Cora.


    It is intuitive that a wider exploration of candidate architectures within a single GNAS task increases our chance to discover better architectures. Thus, we count the average number of unique architectures found by various LLMs during the GNAS iterations, as in Figure~\ref{fig:iso_perf}b.

     As shown in Figure~\ref{fig:iso_perf}b, GPT-4 discovers the most unique architectures across all the four datasets, with an average of 123.8 unique architectures. It also achieves the best performance across all the four datasets. In contrast, ChatGLM3-6B discovers fewer number of unique architectures, only 9.0 across the four datasets, and thus exhibits the worst performance. 
     On the Citeseer dataset, GPT-3.5-turbo discovers a larger number of unique architectures and exhibits better results than PaLM. However, on the arXiv dataset, PaLM 2 wins with 67.3 unique architectures and 72.0\% accuracy.
     It is evident that the number of unique architectures found by LLMs during multiple rounds of search impacts the performance of the LLMs. LLMs that explore more unique architectures generally have a higher chance to find a better  GNN.

    The case study  demonstrates that  our method can be extended to various LLMs. In order to evaluate the inference capability of different LLMs on GNAS tasks, we use the number of unique architectures evaluated by LLMs during the architecture search as an evaluation metric.

\subsection{Case study 3: Comparing in heterogeneous graph for link prediction.}
To investigate whether our method can be generalized to  heterogeneous graphs and new learning tasks, we further conducted link prediction on heterogeneous graphs. Specifically, we continued to use the search space and experimental settings of AutoGEL and conducted searches on the datasets FB15k-237~\cite{toutanova2015observed} and WN18RR~\cite{dettmers2018convolutional}. As the results shown in Table~\ref{tab:AUTOGEL_lp} . 
Our method outperforms AutoGEL in terms of  all the four metrics on the two datasets, demonstrating the good generalization of our method. In terms of the MRR metric, GNAS-LLM is on average 0.42\% ahead. Meanwhile, in the Hit@N metric where N is 10, 3, and 1 respectively, our method is on average most ahead in Hit@1, reaching 0.49\%.

The case study on heterogeneous graph link prediction which can further show GNAS-LLM's potential applicability to a wide range of GNN design problems.

\begin{table*}[h!]
\centering
    \caption{Results of LP task on knowledge graphs \textit{w.r.t.} MRR and Hits@N. The best results are in bold.}
    \label{tab:AUTOGEL_lp}
    \fontsize{8}{1\baselineskip}\selectfont
    \setlength{\tabcolsep}{3pt}
    \begin{tabular}{lcccccccc}
        \toprule
        \multirow{2}{*}{Model}& \multicolumn{4}{c}{FB15k-237} & \multicolumn{4}{c}{WN18RR}
         \\  \cmidrule(lr){2-5}\cmidrule(lr){6-9}
         &MRR & Hits@10  & Hits@3 & Hits@1 & MRR & Hits@10  & Hits@3 & Hits@1    \\
        \midrule 
        AutoGEL&0.3518&0.5305&0.3846&0.2626&0.4630& 0.5255&0.4761 &0.4293\\
        GNAS-LLM&\textbf{0.3539}&\textbf{0.5319}&\textbf{0.3873}&\textbf{0.2646}&\textbf{0.4693}&\textbf{0.5281}&\textbf{0.4818}&\textbf{0.4371}\\
        \bottomrule
    \end{tabular}
\end{table*}

\subsection{Case study 4: Comparing on a large graph}
We conduct experiment on a large graph dataset, ogbn-products~\cite{hu2020open}, which contains 2,449,029 nodes and 61,859,140 edges. Each node has 100 features, and a total of 47 different node classes.
 We use the functions provided by the SGL library~\cite{zhang2022pasca}. For the search process, we use GNAS-LLM to search 10 architectures in each of the 15 rounds. We also use random search and Pasca~\cite{zhang2022pasca} as baselines, exploring a total of 150 architectures. Ultimately, our method achieves an accuracy of 70.86\% on the test set, which is 0.31\% higher than 70.55\% achieved by the random search and 1.36\% higher than 69.50\% achieved by the Pasca.
 
 The case study, we compared our approach with the baseline in the search space specifically designed for large graphs, and our method achieved leading results. This demonstrates that our approach can also handle the architecture search of large-scale graphs.

\section{Discussions}
Based on the experiments, we have the following observations and discussions:

\underline{\textbf{Observation 1}}. \textbf{LLMs are capable of finding the best graph neural architectures by using the GNAS prompts.} As shown in Table~\ref{tab:labelexperiment1}, GNAS-LLM performs better than the baseline methods on average, indicating that LLMs indeed understand the graph architecture search task and are capable of using the GNAS prompts for learning. To sum up, LLMs are capable of running  architecture search on graphs under the GNAS prompts.

\underline{\textbf{Observation 2}}. \textbf{The adjacency matrix outperforms edge lists in guiding LLMs towards generating accurate GNN  architectures.}
As shown in Table~\ref{tab:experiment4},  the variant of GNAS-LLM ``with $Tuple$", which describes model architecture using triplets, underperforms GNAS-LLM by 0.75\% on average across four datasets. This indicates that the adjacency matrix and operation list help LLMs understanding GNN architectures. 
Recent effort to employing LLMs for graph analysis tasks use edge lists as input show restricted performance~\cite{wang2024can}. Nevertheless, while the adjacency matrix format proves effective, it consumes plenty of tokens within LLMs. 

\underline{\textbf{Observation 3}}. \textbf{GPT-4 explores the largest number of unique graph neural architectures by reducing the number of repeated neural architectures.}
Figure~\ref{fig:iso_perf}b shows the number of unique graph neural architectures designed by different LLMs during the search process, with GPT-4 exploring the most unique architectures across all the datasets. In Figure~\ref{fig:iso_perf}a, GPT-4 also achieves the best results on the four datasets. 
These results indicate that with the largest number of parameters amongst the four LLMs, GPT-4 performs the best on the GNAS task, by avoiding  unnecessary computations on the architectures that are different  but close.

\underline{\textbf{Observation 4}}. \textbf{GNAS-LLM outperforms traditional RL-based and evolution-based GNAS methods by running fewer times of iterations, which shows fast convergence of GNAS-LLM.} The experiments demonstrate that our method is capable of identifying the best model on the given validation set. As shown in Figure~\ref{fig:200times}, GNAS-LLM, which uses only 15 iterations (including 150 GNNs evaluated), outperforms RL-based and genetic-based GNAS methods with 200 training iterations (including 2,000 GNNs evaluated) on Cora, Citeseer, and Arxiv. Therefore, it is a good solution to use LLMs for the complicated GNAS tasks.

\underline{\textbf{Observation 5}}. \textbf{LLMs are sensitive to the search strategy, where a well-designed search strategy prompt is important.}
As shown in Table~\ref{tab:experiment4}, the variant of ``$\neg Strategy$", which removes the search strategy from the GNAS prompts, causing the ranking drop of the generated architectures from 1.75 to 23.50. 
The variant ``with $Evolutionary$"  which uses the evolutionary algorithm as search strategy also underperforms GNAS-LLM with an average ranking drop about 43.67 on the datasets.
This result indicates that a well-designed search strategy for LLMs are helpful. 
Furthermore, on the Cora, Pubmed, and arXiv datasets, replacing the GNAS-LLM search strategy with evolutionary algorithm leads to worse results than removing the GNAS-LLM 
 search strategy. This indicates that LLMs are sensitive to the search strategy~\cite{wang2024can}.

\section{Conclusions}
In this paper, we present a new graph neural architecture search method based on Large Language Models, namely GNAS-LLM. We design a new class of \textit{GNAS prompts} that enable LLMs to understand existing graph search space, search strategy, and search feedback. By leveraging the powerful generative capability of LLMs, GNAS-LLM  generates better GNN architectures than existing GNAS methods. 
Experimental results show that GNAS-LLM can generate the best GNN architectures by exploring 97.5 GNN architectures on average on the benchmark datasets. Moreover, ablation studies show that the GNAS prompts  are successful in generating accurate architectures. Last but not least, the reward of the GNAS prompts promotes fast convergence of GNAS-LLM by continuously fine-tuning the prompts based on the model evaluation feedback. 

In the future, we will test the robustness and adaptability of GNAS-LLM when meeting a broader range of GNN architectures. 
It is also interesting yet challenging to embed LLMs into heterogeneous  graph neural architecture search where the search space and search strategy are more complicated. 
The method is still rely on computationally intensive LLMs, which limits their use in low-resource settings. Therefore, developing a computation-efficient GNAS-LLM is also valuable.

\Acknowledgements{This work is supported by the National Natural Science Foundation of China (Grant Nos. 62406279, 62202422, and 62376064), Natural Science Foundation of Shandong Province (Grant No. ZR2021MH227), and Shanghai Artificial Intelligence Laboratory.}

\bibliographystyle{scis}
\bibliography{ref}

\begin{thebibliography}{10}
\providecommand{\url}[1]{\texttt{#1}}
\providecommand{\urlprefix}{URL }
\providecommand{\bibinfo}[2]{#2}

\bibitem{wang2016incremental}
\bibinfo{author}{Wang H}, \bibinfo{author}{Zhang P}, \bibinfo{author}{Zhu X},
  et~al.
\newblock \bibinfo{title}{Incremental subgraph feature selection for graph
  classification}.
\newblock \bibinfo{journal}{IEEE Transactions on Knowledge and Data
  Engineering}, \bibinfo{year}{2016}, \bibinfo{volume}{29}:
  \bibinfo{pages}{128--142}

\bibitem{gaoprecision}
\bibinfo{author}{Gao Y}, \bibinfo{author}{Zhang X}, \bibinfo{author}{Sun Z},
  et~al.
\newblock \bibinfo{title}{Precision adverse drug reactions prediction with
  heterogeneous graph neural network}.
\newblock \bibinfo{journal}{Advanced science (Weinheim, Baden-Wurttemberg,
  Germany)}: \bibinfo{pages}{e2404671}

\bibitem{chen2024deepasd}
\bibinfo{author}{Chen W}, \bibinfo{author}{Yang J}, \bibinfo{author}{Sun Z},
  et~al.
\newblock \bibinfo{title}{Deepasd: a deep adversarial-regularized graph
  learning method for asd diagnosis with multimodal data}.
\newblock \bibinfo{journal}{Translational Psychiatry}, \bibinfo{year}{2024},
  \bibinfo{volume}{14}: \bibinfo{pages}{375}

\bibitem{li2021live}
\bibinfo{author}{Li Z}, \bibinfo{author}{Wang H}, \bibinfo{author}{Zhang P},
  et~al.
\newblock \bibinfo{title}{Live-streaming fraud detection: A heterogeneous graph
  neural network approach}.
\newblock In: \bibinfo{booktitle}{Proceedings of the 27th ACM SIGKDD Conference
  on Knowledge Discovery \& Data Mining}, \bibinfo{year}{2021}.
\newblock \bibinfo{pages}{3670--3678}

\bibitem{DBLP:conf/ijcai/ZhangW021}
\bibinfo{author}{Zhang Z}, \bibinfo{author}{Wang X}, \bibinfo{author}{Zhu W}.
\newblock \bibinfo{title}{Automated machine learning on graphs: {A} survey}.
\newblock In: \bibinfo{editor}{Z~Zhou}, ed., \bibinfo{booktitle}{Proceedings of
  the Thirtieth International Joint Conference on Artificial Intelligence,
  {IJCAI} 2021, Virtual Event / Montreal, Canada, 19-27 August 2021},
  \bibinfo{year}{2021}.
\newblock \bibinfo{pages}{4704--4712}

\bibitem{DBLP:conf/ijcai/GaoYZ0H20}
\bibinfo{author}{Gao Y}, \bibinfo{author}{Yang H}, \bibinfo{author}{Zhang P},
  et~al.
\newblock \bibinfo{title}{Graph neural architecture search}.
\newblock In: \bibinfo{booktitle}{Proc. of IJCAI}, \bibinfo{year}{2020}.
\newblock \bibinfo{pages}{1403--1409}

\bibitem{DBLP:conf/icde/ZhaoYT21}
\bibinfo{author}{Huan Z}, \bibinfo{author}{Quanming Y}, \bibinfo{author}{Weiwei
  T}.
\newblock \bibinfo{title}{Search to aggregate neighborhood for graph neural
  network}.
\newblock In: \bibinfo{booktitle}{2021 IEEE 37th International Conference on
  Data Engineering (ICDE)}, \bibinfo{year}{2021}.
\newblock \bibinfo{pages}{552--563}

\bibitem{Shi2020EvolutionaryAS}
\bibinfo{author}{Shi M}, \bibinfo{author}{Tang Y}, \bibinfo{author}{Zhu X},
  et~al.
\newblock \bibinfo{title}{Genetic-gnn: Evolutionary architecture search for
  graph neural networks}.
\newblock \bibinfo{journal}{Knowledge-Based Systems}, \bibinfo{year}{2022}

\bibitem{zheng2023can}
\bibinfo{author}{Zheng M}, \bibinfo{author}{Su X}, \bibinfo{author}{You S},
  et~al.
\newblock \bibinfo{title}{Can gpt-4 perform neural architecture search?}
\newblock \bibinfo{journal}{ArXiv preprint}, \bibinfo{year}{2023}

\bibitem{DBLP:conf/cvpr/Cai0DZZ0H21}
\bibinfo{author}{Cai S}, \bibinfo{author}{Li L}, \bibinfo{author}{Deng J},
  et~al.
\newblock \bibinfo{title}{Rethinking graph neural architecture search from
  message-passing}.
\newblock In: \bibinfo{booktitle}{Proc. of CVPR}, \bibinfo{year}{2021}.
\newblock \bibinfo{pages}{6657--6666}

\bibitem{zhang2023autogt}
\bibinfo{author}{Zhang Z}, \bibinfo{author}{Wang X}, \bibinfo{author}{Guan C},
  et~al.
\newblock \bibinfo{title}{Auto{GT}: Automated graph transformer architecture
  search}.
\newblock In: \bibinfo{booktitle}{The Eleventh International Conference on
  Learning Representations}, \bibinfo{year}{2023}.
\newblock \urlprefix\url{https://openreview.net/forum?id=GcM7qfl5zY}

\bibitem{AutoGNN}
\bibinfo{author}{Zhou K}, \bibinfo{author}{Huang X}, \bibinfo{author}{Song Q},
  et~al.
\newblock \bibinfo{title}{Auto-gnn: Neural architecture search of graph neural
  networks}.
\newblock \bibinfo{journal}{Frontiers in big Data}, \bibinfo{year}{2022}

\bibitem{gao2022graphnas++}
\bibinfo{author}{Gao Y}, \bibinfo{author}{Zhang P}, \bibinfo{author}{Yang H},
  et~al.
\newblock \bibinfo{title}{Graphnas++: Distributed architecture search for graph
  neural networks}.
\newblock \bibinfo{journal}{IEEE Transactions on Knowledge and Data
  Engineering}, \bibinfo{year}{2022}

\bibitem{gao2023gm2nas}
\bibinfo{author}{Gao J}, \bibinfo{author}{Al-Sabri R},
  \bibinfo{author}{Oloulade B~M}, et~al.
\newblock \bibinfo{title}{Gm2nas: multitask multiview graph neural architecture
  search}.
\newblock \bibinfo{journal}{Knowledge and Information Systems},
  \bibinfo{year}{2023}, \bibinfo{volume}{65}: \bibinfo{pages}{4021--4054}

\bibitem{al2022multi}
\bibinfo{author}{Al-Sabri R}, \bibinfo{author}{Gao J}, \bibinfo{author}{Chen
  J}, et~al.
\newblock \bibinfo{title}{Multi-view graph neural architecture search for
  biomedical entity and relation extraction}.
\newblock \bibinfo{journal}{IEEE/ACM Transactions on Computational Biology and
  Bioinformatics}, \bibinfo{year}{2022}, \bibinfo{volume}{20}:
  \bibinfo{pages}{1221--1233}

\bibitem{DBLP:conf/icdm/GaoZLZLH21}
\bibinfo{author}{Gao Y}, \bibinfo{author}{Zhang P}, \bibinfo{author}{Li Z},
  et~al.
\newblock \bibinfo{title}{Heterogeneous graph neural architecture search}.
\newblock In: \bibinfo{booktitle}{{IEEE} International Conference on Data
  Mining, {ICDM} 2021, Auckland, New Zealand, December 7-10, 2021},
  \bibinfo{year}{2021}.
\newblock \bibinfo{pages}{1066--1071}

\bibitem{gao2023hgnas++}
\bibinfo{author}{Gao Y}, \bibinfo{author}{Zhang P}, \bibinfo{author}{Zhou C},
  et~al.
\newblock \bibinfo{title}{Hgnas++: efficient architecture search for
  heterogeneous graph neural networks}.
\newblock \bibinfo{journal}{IEEE Transactions on Knowledge and Data
  Engineering}, \bibinfo{year}{2023}

\bibitem{li2021one}
\bibinfo{author}{Li Y}, \bibinfo{author}{Wen Z}, \bibinfo{author}{Wang Y},
  et~al.
\newblock \bibinfo{title}{One-shot graph neural architecture search with
  dynamic search space}.
\newblock In: \bibinfo{booktitle}{Proc. of AAAI}, \bibinfo{year}{2021}.
\newblock \bibinfo{pages}{8510--8517}

\bibitem{guan2022large}
\bibinfo{author}{Guan C}, \bibinfo{author}{Wang X}, \bibinfo{author}{Chen H},
  et~al.
\newblock \bibinfo{title}{Large-scale graph neural architecture search}.
\newblock In: \bibinfo{booktitle}{Proc. of ICML}, \bibinfo{year}{2022}.
\newblock \bibinfo{pages}{7968--7981}

\bibitem{qin2022graph}
\bibinfo{author}{Qin Y}, \bibinfo{author}{Wang X}, \bibinfo{author}{Zhang Z},
  et~al.
\newblock \bibinfo{title}{Graph neural architecture search under distribution
  shifts}.
\newblock In: \bibinfo{booktitle}{Proc. of ICML}, \bibinfo{year}{2022}.
\newblock \bibinfo{pages}{18083--18095}

\bibitem{zheng2023auto}
\bibinfo{author}{Zheng X}, \bibinfo{author}{Zhang M}, \bibinfo{author}{Chen C},
  et~al.
\newblock \bibinfo{title}{Auto-heg: Automated graph neural network on
  heterophilic graphs}.
\newblock In: \bibinfo{booktitle}{Proceedings of the {ACM} Web Conference 2023,
  {WWW} 2023, Austin, TX, USA, 30 April 2023 - 4 May 2023},
  \bibinfo{year}{2023}.
\newblock \bibinfo{pages}{611--620}

\bibitem{wang2021autogel}
\bibinfo{author}{Wang Z}, \bibinfo{author}{Di S}, \bibinfo{author}{Chen L}.
\newblock \bibinfo{title}{Autogel: An automated graph neural network with
  explicit link information}.
\newblock In: \bibinfo{booktitle}{Advances in Neural Information Processing
  Systems 34: Annual Conference on Neural Information Processing Systems 2021,
  NeurIPS 2021, December 6-14, 2021, virtual}, \bibinfo{year}{2021}.
\newblock \bibinfo{pages}{24509--24522}

\bibitem{diffmg}
\bibinfo{author}{Ding Y}, \bibinfo{author}{Yao Q}, \bibinfo{author}{Zhao H},
  et~al.
\newblock \bibinfo{title}{Diffmg: Differentiable meta graph search for
  heterogeneous graph neural networks}.
\newblock In: \bibinfo{booktitle}{{KDD} '21: The 27th {ACM} {SIGKDD} Conference
  on Knowledge Discovery and Data Mining, Virtual Event, Singapore, August
  14-18, 2021}, \bibinfo{year}{2021}.
\newblock \bibinfo{pages}{279--288}

\bibitem{Zheng2022MultiRelationalGN}
\bibinfo{author}{Zheng X}, \bibinfo{author}{Zhang M}, \bibinfo{author}{cheng
  Jason~Chen C}, et~al.
\newblock \bibinfo{title}{Multi-relational graph neural architecture search
  with fine-grained message passing}.
\newblock \bibinfo{journal}{2022 IEEE International Conference on Data Mining
  (ICDM)}, \bibinfo{year}{2022}

\bibitem{Zhang_Zhang_Wang_Qin_Qin_Zhu_2023}
\bibinfo{author}{Zhang Z}, \bibinfo{author}{Zhang Z}, \bibinfo{author}{Wang X},
  et~al.
\newblock \bibinfo{title}{Dynamic heterogeneous graph attention neural
  architecture search}.
\newblock In: \bibinfo{booktitle}{Thirty-Seventh {AAAI} Conference on
  Artificial Intelligence, {AAAI} 2023, Thirty-Fifth Conference on Innovative
  Applications of Artificial Intelligence, {IAAI} 2023, Thirteenth Symposium on
  Educational Advances in Artificial Intelligence, {EAAI} 2023, Washington, DC,
  USA, February 7-14, 2023}, \bibinfo{year}{2023}.
\newblock \bibinfo{pages}{11307--11315}

\bibitem{zhang2024meta}
\bibinfo{author}{Zhang X}, \bibinfo{author}{Gao Y}, \bibinfo{author}{Liu Y},
  et~al.
\newblock \bibinfo{title}{Meta structure search for link weight prediction in
  heterogeneous graphs}.
\newblock In: \bibinfo{booktitle}{ICASSP 2024-2024 IEEE International
  Conference on Acoustics, Speech and Signal Processing (ICASSP)},
  \bibinfo{year}{2024}.
\newblock \bibinfo{pages}{5195--5199}

\bibitem{li2020autograph}
\bibinfo{author}{Li Y}, \bibinfo{author}{King I}.
\newblock \bibinfo{title}{Autograph: Automated graph neural network}.
\newblock In: \bibinfo{booktitle}{Neural Information Processing - 27th
  International Conference, {ICONIP} 2020, Bangkok, Thailand, November 23-27,
  2020, Proceedings, Part {II}}, \bibinfo{year}{2020},
  \emph{\bibinfo{series}{Lecture Notes in Computer Science}}, volume
  \bibinfo{volume}{12533}.
\newblock \bibinfo{pages}{189--201}

\bibitem{DBLP:conf/cvpr/XieCZWWZY023}
\bibinfo{author}{Xie B}, \bibinfo{author}{Chang H}, \bibinfo{author}{Zhang Z},
  et~al.
\newblock \bibinfo{title}{Adversarially robust neural architecture search for
  graph neural networks}.
\newblock In: \bibinfo{booktitle}{{IEEE/CVF} Conference on Computer Vision and
  Pattern Recognition, {CVPR} 2023, Vancouver, BC, Canada, June 17-24, 2023},
  \bibinfo{year}{2023}.
\newblock \bibinfo{pages}{8143--8152}

\bibitem{oloulade2021graph}
\bibinfo{author}{Oloulade B~M}, \bibinfo{author}{Gao J}, \bibinfo{author}{Chen
  J}, et~al.
\newblock \bibinfo{title}{Graph neural architecture search: A survey}.
\newblock \bibinfo{journal}{Tsinghua Science and Technology},
  \bibinfo{year}{2021}

\bibitem{openai2023gpt4}
\bibinfo{author}{Achiam J}, \bibinfo{author}{Adler S}, \bibinfo{author}{Agarwal
  S}, et~al.
\newblock \bibinfo{title}{Gpt-4 technical report}.
\newblock \bibinfo{journal}{arXiv preprint arXiv:2303.08774},
  \bibinfo{year}{2023}

\bibitem{guo2023gpt4graph}
\bibinfo{author}{Guo J}, \bibinfo{author}{Du L}, \bibinfo{author}{Liu H}.
\newblock \bibinfo{title}{Gpt4graph: Can large language models understand graph
  structured data? an empirical evaluation and benchmarking}.
\newblock \bibinfo{journal}{ArXiv preprint}, \bibinfo{year}{2023}

\bibitem{zhanggraph}
\bibinfo{author}{Zhang J}.
\newblock \bibinfo{title}{Graph-toolformer: To empower llms with graph
  reasoning ability via prompt augmented by chatgpt}.
\newblock \bibinfo{journal}{ArXiv preprint}, \bibinfo{year}{2023}

\bibitem{DBLP:conf/naacl/DevlinCLT19}
\bibinfo{author}{Devlin J}, \bibinfo{author}{Chang M~W}, \bibinfo{author}{Lee
  K}, et~al.
\newblock \bibinfo{title}{{BERT}: Pre-training of deep bidirectional
  transformers for language understanding}.
\newblock In: \bibinfo{booktitle}{Proceedings of the 2019 Conference of the
  North American Chapter of the Association for Computational Linguistics:
  Human Language Technologies, {NAACL-HLT} 2019, Minneapolis, MN, USA, June
  2-7, 2019, Volume 1 (Long and Short Papers)}, \bibinfo{year}{2019}.
\newblock \bibinfo{pages}{4171--4186}

\bibitem{Lan2020ALBERT}
\bibinfo{author}{Lan Z}, \bibinfo{author}{Chen M}, \bibinfo{author}{Goodman S},
  et~al.
\newblock \bibinfo{title}{Albert: A lite bert for self-supervised learning of
  language representations}.
\newblock In: \bibinfo{booktitle}{International Conference on Learning
  Representations}, \bibinfo{year}{2020}.
\newblock \urlprefix\url{https://openreview.net/forum?id=H1eA7AEtvS}

\bibitem{DBLP:conf/iclr/HeLGC21}
\bibinfo{author}{He P}, \bibinfo{author}{Liu X}, \bibinfo{author}{Gao J},
  et~al.
\newblock \bibinfo{title}{Deberta: decoding-enhanced bert with disentangled
  attention}.
\newblock In: \bibinfo{booktitle}{Proc. of ICLR}, \bibinfo{year}{2021}

\bibitem{chowdhery2022palm}
\bibinfo{author}{Chowdhery A}, \bibinfo{author}{Narang S},
  \bibinfo{author}{Devlin J}, et~al.
\newblock \bibinfo{title}{Palm: Scaling language modeling with pathways}.
\newblock \bibinfo{journal}{arXiv preprint arXiv:2204.02311},
  \bibinfo{year}{2022}

\bibitem{Vaswani2017AttentionIA}
\bibinfo{author}{Vaswani A}, \bibinfo{author}{Shazeer N},
  \bibinfo{author}{Parmar N}, et~al.
\newblock \bibinfo{title}{Attention is all you need}.
\newblock In: \bibinfo{booktitle}{Advances in Neural Information Processing
  Systems 30: Annual Conference on Neural Information Processing Systems 2017,
  December 4-9, 2017, Long Beach, CA, {USA}}, \bibinfo{year}{2017}.
\newblock \bibinfo{pages}{5998--6008}

\bibitem{anil2023palm}
\bibinfo{author}{Anil R}, \bibinfo{author}{Dai A~M}, \bibinfo{author}{Firat O},
  et~al.
\newblock \bibinfo{title}{Palm 2 technical report}.
\newblock \bibinfo{journal}{ArXiv preprint}, \bibinfo{year}{2023}

\bibitem{DBLP:journals/corr/abs-2311-12399}
\bibinfo{author}{Li Y}, \bibinfo{author}{Li Z}, \bibinfo{author}{Wang P},
  et~al.
\newblock \bibinfo{title}{A survey of graph meets large language model:
  Progress and future directions}.
\newblock \bibinfo{journal}{CoRR}, \bibinfo{year}{2023},
  \bibinfo{volume}{abs/2311.12399}

\bibitem{he2023harnessing}
\bibinfo{author}{He X}, \bibinfo{author}{Bresson X}, \bibinfo{author}{Laurent
  T}, et~al.
\newblock \bibinfo{title}{Harnessing explanations: Llm-to-lm interpreter for
  enhanced text-attributed graph representation learning}, \bibinfo{year}{2023}

\bibitem{tang2024graphgptgraphinstructiontuning}
\bibinfo{author}{Tang J}, \bibinfo{author}{Yang Y}, \bibinfo{author}{Wei W},
  et~al.
\newblock \bibinfo{title}{Graphgpt: Graph instruction tuning for large language
  models}, \bibinfo{year}{2024}

\bibitem{canwesoftpromptllms}
\bibinfo{author}{Liu Z}, \bibinfo{author}{He X}, \bibinfo{author}{Tian Y},
  et~al.
\newblock \bibinfo{title}{Can we soft prompt llms for graph learning tasks?}
\newblock In: \bibinfo{booktitle}{Companion Proceedings of the {ACM} on Web
  Conference 2024, {WWW} 2024, Singapore, Singapore, May 13-17, 2024},
  \bibinfo{year}{2024}.
\newblock \bibinfo{pages}{481--484}

\bibitem{zhang2023automl}
\bibinfo{author}{Zhang S}, \bibinfo{author}{Gong C}, \bibinfo{author}{Wu L},
  et~al.
\newblock \bibinfo{title}{Automl-gpt: Automatic machine learning with gpt}.
\newblock \bibinfo{journal}{ArXiv preprint}, \bibinfo{year}{2023}

\bibitem{qin2022bench}
\bibinfo{author}{Qin Y}, \bibinfo{author}{Zhang Z}, \bibinfo{author}{Wang X},
  et~al.
\newblock \bibinfo{title}{Nas-bench-graph: Benchmarking graph neural
  architecture search}.
\newblock \bibinfo{journal}{Proc. of NeurIPS}, \bibinfo{year}{2022}

\bibitem{zhang2023llm4dyg}
\bibinfo{author}{Zhang Z}, \bibinfo{author}{Wang X}, \bibinfo{author}{Zhang Z},
  et~al.
\newblock \bibinfo{title}{Llm4dyg: Can large language models solve problems on
  dynamic graphs?}
\newblock \bibinfo{journal}{arXiv preprint arXiv:2310.17110},
  \bibinfo{year}{2023}

\bibitem{huang2023can}
\bibinfo{author}{Huang J}, \bibinfo{author}{Zhang X}, \bibinfo{author}{Mei Q},
  et~al.
\newblock \bibinfo{title}{Can llms effectively leverage graph structural
  information: when and why}.
\newblock \bibinfo{journal}{arXiv preprint arXiv:2309.16595},
  \bibinfo{year}{2023}

\bibitem{wang2024can}
\bibinfo{author}{Wang H}, \bibinfo{author}{Feng S}, \bibinfo{author}{He T},
  et~al.
\newblock \bibinfo{title}{Can language models solve graph problems in natural
  language?}
\newblock \bibinfo{journal}{Advances in Neural Information Processing Systems},
  \bibinfo{year}{2024}, \bibinfo{volume}{36}

\bibitem{toutanova2015observed}
\bibinfo{author}{Toutanova K}, \bibinfo{author}{Chen D}.
\newblock \bibinfo{title}{Observed versus latent features for knowledge base
  and text inference}.
\newblock In: \bibinfo{booktitle}{Proceedings of the 3rd workshop on continuous
  vector space models and their compositionality}, \bibinfo{year}{2015}.
\newblock \bibinfo{pages}{57--66}

\bibitem{dettmers2018convolutional}
\bibinfo{author}{Dettmers T}, \bibinfo{author}{Minervini P},
  \bibinfo{author}{Stenetorp P}, et~al.
\newblock \bibinfo{title}{Convolutional 2d knowledge graph embeddings}.
\newblock In: \bibinfo{booktitle}{Proceedings of the Thirty-Second {AAAI}
  Conference on Artificial Intelligence, (AAAI-18), the 30th innovative
  Applications of Artificial Intelligence (IAAI-18), and the 8th {AAAI}
  Symposium on Educational Advances in Artificial Intelligence (EAAI-18), New
  Orleans, Louisiana, USA, February 2-7, 2018}, \bibinfo{year}{2018}.
\newblock \bibinfo{pages}{1811--1818}

\bibitem{zeng2022glm}
\bibinfo{author}{Zeng A}, \bibinfo{author}{Liu X}, \bibinfo{author}{Du Z},
  et~al.
\newblock \bibinfo{title}{Glm-130b: An open bilingual pre-trained model}.
\newblock \bibinfo{journal}{ArXiv preprint}, \bibinfo{year}{2022}

\bibitem{brown2020language}
\bibinfo{author}{Brown T}, \bibinfo{author}{Mann B}, \bibinfo{author}{Ryder N},
  et~al.
\newblock \bibinfo{title}{Language models are few-shot learners}.
\newblock \bibinfo{journal}{Advances in neural information processing systems},
  \bibinfo{year}{2020}, \bibinfo{volume}{33}: \bibinfo{pages}{1877--1901}

\bibitem{hu2020open}
\bibinfo{author}{Hu W}, \bibinfo{author}{Fey M}, \bibinfo{author}{Zitnik M},
  et~al.
\newblock \bibinfo{title}{Open graph benchmark: Datasets for machine learning
  on graphs}.
\newblock \bibinfo{journal}{Advances in neural information processing systems},
  \bibinfo{year}{2020}, \bibinfo{volume}{33}: \bibinfo{pages}{22118--22133}

\bibitem{zhang2022pasca}
\bibinfo{author}{Zhang W}, \bibinfo{author}{Shen Y}, \bibinfo{author}{Lin Z},
  et~al.
\newblock \bibinfo{title}{Pasca: A graph neural architecture search system
  under the scalable paradigm}.
\newblock In: \bibinfo{booktitle}{{WWW} '22: The {ACM} Web Conference 2022,
  Virtual Event, Lyon, France, April 25 - 29, 2022}, \bibinfo{year}{2022}.
\newblock \bibinfo{pages}{1817--1828}

\end{thebibliography}

\end{document}